\RequirePackage{fix-cm}

\documentclass[twocolumn]{article}

\usepackage{natbib}

\usepackage[utf8]{inputenc}
\usepackage{times}
\usepackage{epsfig}
\usepackage{epstopdf}
\usepackage{graphicx}
\usepackage{amsmath}
\usepackage{amssymb}
\usepackage{color,soul}
\usepackage{algorithm}
\usepackage{algorithmic}
\usepackage{multirow}
\usepackage{tikz}
\usepackage{url}

\usepackage{comment}
\usepackage[misc]{ifsym}

\usepackage{subcaption}
\usepackage{hyperref}

\graphicspath{{./figures/}}

\definecolor{gold}{HTML}{EEAD0E}
\definecolor{silver}{HTML}{C0C0C0}
\definecolor{bronze}{HTML}{CD7F32}
           
\newcommand{\first}[1]{%
    {\raisebox{0.8pt}{\footnotesize \color{gold} \circledd{1}}\hspace{3.5pt}#1}%
}
\newcommand{\second}[1]{%
    {\raisebox{0.8pt}{\footnotesize \color{silver} \circledd{2}}\hspace{3.5pt}#1}%
}
\newcommand{\third}[1]{%
    {\raisebox{0.8pt}{\footnotesize \color{bronze} \circledd{3}}\hspace{3.5pt}#1}%
}

\def\imagebox#1#2{\vtop to #1{\null\hbox{#2}\vfill}}

\newcommand{\cmark}{x}
\newcommand{\xmark}{--}

\providecommand{\keywords}[1]{\textbf{\textit{Keywords---}} #1}

\date{\vspace{-5ex}}

\begin{document}

\title{Discriminative Correlation Filter Tracker with \\ Channel and Spatial Reliability}  
 


\author{Alan Lukežič$^1$, Tomáš Vojíř$^2$, Luka Čehovin Zajc$^1$, Jiří Matas$^2$ and Matej Kristan$^1$\\
$^1$Faculty of Computer and Information Science, University of Ljubljana, Slovenia \\
$^2$Faculty of Electrical Engineering, Czech Technical University in Prague, Czech Republic \\
{\tt\small \{alan.lukezic, luka.cehovin, matej.kristan\}@fri.uni-lj.si} \\
{\tt\small \{vojirtom, matas\}@cmp.felk.cvut.cz} \\
}





\maketitle

\begin{abstract}
{\it Short-term tracking is an open and challenging  problem for which discriminative correlation filters (DCF) have shown  excellent performance. We introduce the channel and spatial reliability concepts to DCF tracking and provide a learning algorithm for its efficient and seamless integration in the filter update and the tracking process. The spatial reliability map adjusts the  filter support to the part of the object suitable for tracking. This both allows to enlarge the search region and  improves tracking of non-rectangular objects.   Reliability scores reflect channel-wise quality of the learned filters and are used as feature weighting coefficients in localization. Experimentally,  with only two simple standard  feature sets, HoGs and Colornames, the novel CSR-DCF method -- DCF with Channel and Spatial Reliability -- achieves state-of-the-art results on VOT 2016, VOT 2015 and OTB100. The CSR-DCF runs close to real-time on a CPU.}
\end{abstract}

\noindent \keywords{Visual tracking, Correlation filters, Channel reliability, Constrained optimization}

\section{Introduction}  \label{sec:introduction}

Short-term, model-free visual object tracking is the problem of continuously localizing a target in a video-sequence given a single example of its appearance. It has received
significant attention of the computer vision community which is reflected in the number of papers published on the topic and the existence of multiple performance evaluation benchmarks~\citep{otb_cvpr2010,kristan_vot2013,kristan_vot2014,kristan_vot2015,kristan_vot_tpami2016,templecolor_tip2015,alov_pami2014,uav_benchmark_eccv2016}. Diverse factors -- occlusion,  illumination change, fast object or camera motion, appearance changes due to rigid or non-rigid deformations and similarity to the background -- make short-term tracking challenging.

\begin{figure}[!t]
\centering
\includegraphics[width=1\linewidth]{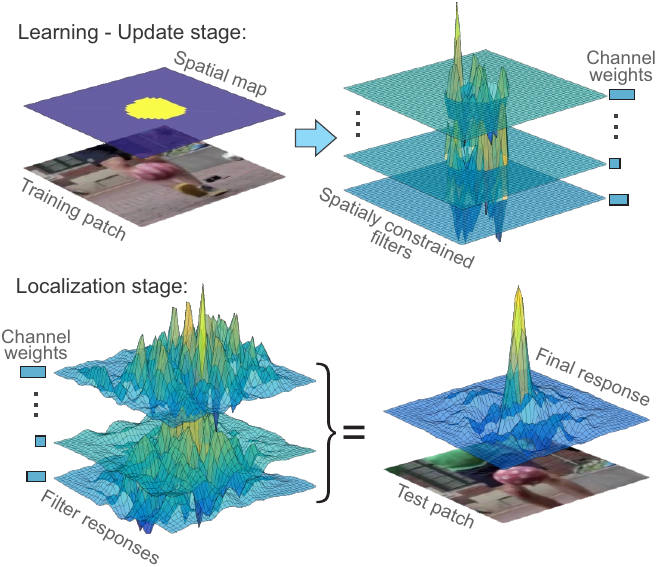}
\caption{Overview of the CSR-DCF approach. An automatically estimated spatial reliability map restricts the correlation filter to the parts suitable for tracking (top) improving localization within a larger search region and performance for irregularly shaped objects. Channel reliability weights calculated in the constrained optimization step of
the correlation filter learning reduce the noise of the weight-averaged filter response~(bottom).}
\label{fig:filter_comparison}
\end{figure}

Recent short-term tracking evaluations ~\citep{otb_cvpr2010,kristan_vot2013,kristan_vot2014,kristan_vot2015} consistently confirm the advantages of semi-supervised discriminative tracking approaches~\citep{grabner_oab,babenko_mil,hare_struck,bolme2010visual}. In particular, trackers based on the discriminative correlation filter (DCF) method~\citep{bolme2010visual,danelljan2014accurate,henriques2015tracking,samf_eccv2014,srdcf_iccv2015} have shown state-of-the-art performance in all standard benchmarks. Discriminative correlation methods learn a filter with a pre-defined response on the training image. The latter is obtained by slightly extending the region around the target to include background samples.

\begin{figure}[!t]
\centering
\includegraphics[width=1\linewidth]{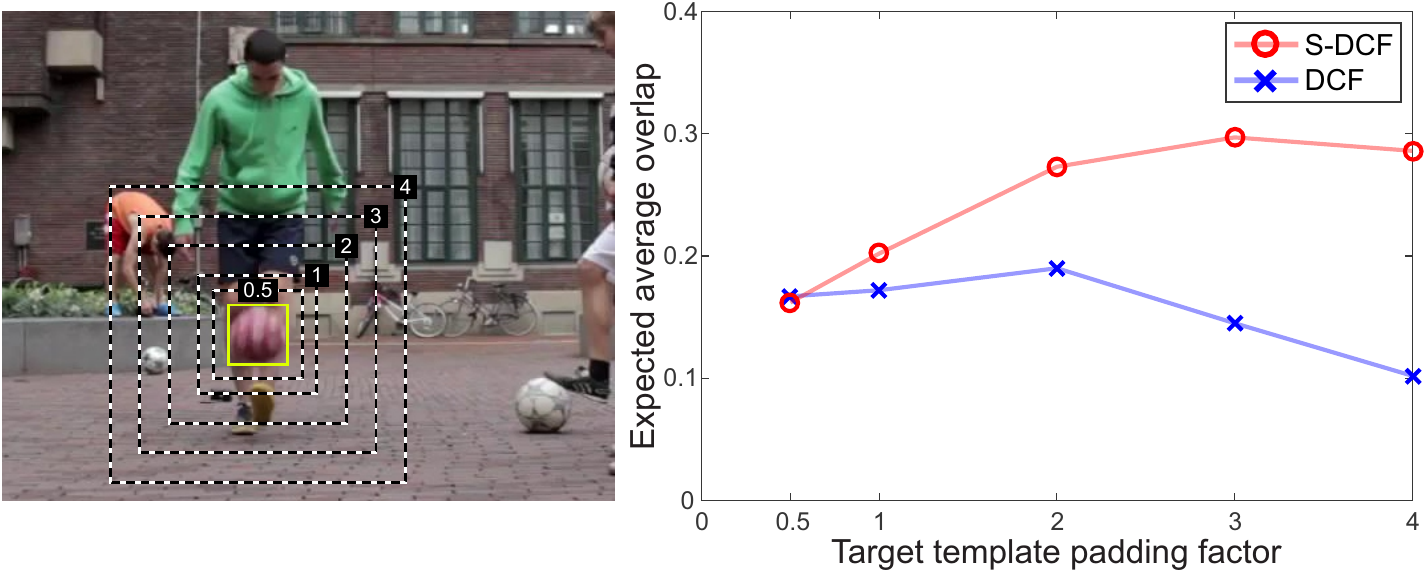}
\caption{Tracking performance measured by the Expected Average Overlap (EAO) of the standard DCF and our spatially constrained DCF (S-DCF) as 
a function of search region size, expressed as 
the multiple of the target size (right, x-axis). The filter is learned from a training region equal in size to the search region. The search region sizes are visualized by black-white dashed rectangles (left image) and the target bounding box is shown in yellow.}
\label{fig:eao_comparison_intro}
\end{figure}

The standard formulation of DCF uses circular correlation which allows to implement learning efficiently by Fast Fourier transform (FFT). However, the FFT requires the filter and the search region size to be equal which limits the detection range. Due to the circularity, the filter is trained on many examples that contain unrealistic, wrapped-around circularly-shifted versions of the target. 
A naive approach to the reduction of the windowing problems is to learn the filter from a larger region. However, due to the large area of the background in the region, the tracking performance of the DCF drops significantly as shown in Figure~\ref{fig:eao_comparison_intro}.

The windowing problems were recently addressed by \cite{cfwlb_cvpr2015} who propose zero-padding the filter during learning and by \cite{srdcf_iccv2015} who introduce spatial regularization to penalize filter values outside the target boundaries. Both approaches train from image regions much larger than the target and thus increase the detection range.

Another limitation of the published DCF methods is the assumption that the target shape is well approximated by an axis-aligned rectangle.  For irregularly shaped objects or those with a hollow center, the filter eventually learns the background, which may lead to drift and failure. The same problem appears for approximately rectangular objects in the case of occlusion. The \cite{cfwlb_cvpr2015} and \cite{srdcf_iccv2015} methods both suffer from this problem.
 
In this paper we introduce the CSR-DCF, the Discriminative Correlation Filter with Channel and Spatial Reliability.  The spatial reliability map adapts the filter support to the part of the object suitable for tracking which overcomes both the problems of circular shift enabling an arbitrary search (and training) region size and the limitations related to the rectangular shape assumption. An important benefit of a large training region is that background samples from a wider area around the target are obtained to improve the filter discriminative power. The spatial reliability map is estimated using the output of a graph labeling problem solved efficiently in each frame. An efficient optimization procedure is applied for learning a correlation filter with the support constrained by the spatial reliability map since the standard closed-form solution cannot be generalized to this case. Figure~\ref{fig:eao_comparison_intro} shows that tracking performance of our spatially constrained correlation filter (denoted as S-DCF) does not degrade with increasing training and search region size as is the case with the standard DCF. In contrast, the performance of S-DCF improves from better treatment of training samples and increased search region size. Experiments show that the novel filter optimization procedure outperforms related approaches for constrained learning in DCFs.

Channel reliability is the second novelty the CSR-DCF tracker introduces. The reliability is estimated from the properties of the constrained least-squares solution to filter design. The channel reliability scores are used for weighting the per-channel filter responses in localization (Figure~\ref{fig:filter_comparison}). 
The CSR-DCF shows state-of-the-art performance on standard benchmarks -- OTB100~\citep{otb_pami2015}, VOT2015~\citep{kristan_vot2015} and  VOT2016~\citep{kristan_vot2015} while running close to real-time on a single CPU. The spatial and channel reliability formulation is general and can be used in most modern correlation filters, e.g. those using deep features. 

The remainder of the paper is structured as follows. In Section~\ref{sec:related_work} we review most closely related work, our approach is described in Section~\ref{sec:approach}, experimental results are presented in Section~\ref{sec:experimental_analysis} and conclusions are drawn in Section~\ref{sec:conclusion}.

\section{Related work}\label{sec:related_work}

The discriminative correlation filters for object detection date back to the 80's with seminal work of \cite{hester1980}. They have been popularized only recently in the tracking community, starting with the \cite{bolme2010visual} MOSSE tracker. Using a gray-scale template, MOSSE achi\-eved  state-of-the-art performance on a tracking benchmark \citep{otb_cvpr2010} at a remarkable processing speed. Significant improvements have been made since and in 2014 the top-performing trackers on a recent benchmark~\citep{kristan_vot2014} were all from this class of trackers.  DCF improvements fall into two categories,  introduction of new features and conceptual improvements in filter learning.

In the first group, \cite{henriques2015tracking} replaced the grayscale templates by HoG~\citep{dalal_triggs_hog}, \cite{danelljan2014adaptive} proposed multi-dimensional color attributes and \cite{Li2014} applied feature combination. Recently, convolutional network features learned for object detection have been applied~\citep{convolutional_mingsungyang_iccv2015, danelljan_iccv2015_convolutional, danelljan_eccv2016_ccot}, leading to a performance boost, but at a cost of significant speed reduction.

Conceptually, the first successful theoretical extension of the standard DCF was the kernelized formulation by \cite{henriques2015tracking} which achieved remarkable tracking performance, but still preserved high speed. Later, a correlation filter based scale adaptation was proposed by \cite{danelljan2014accurate} introduced a scale-space pyramid learned within a correlation filter framework. \cite{zhang_stc_eccv2014} introduced spatio-temporal context learning in the DCFs.
To improve localization with correlation filters, \cite{staple_cvpr2016} proposed a tracking method that combines the output of the correlation filter with the target segmentation probability map. \cite{danelljan_eccv2016_ccot} addressed a multiple-resolution feature map issue in correlation filters by formulating filter learning in continuous space, while \cite{qi_hedge_deep_cf} proposed a mechanism to combine correlation responses from multiple convolutional layers. A correlation filter tracker which is able to handle drifts in longer sequences was proposed by \cite{wang_reliable_memories}. It clusters similar target appearances together and uses the clusters for target localization instead of a single online learned filter. 

Since most of the correlation filter trackers represent the target with a single filter, it can easily get corrupted when occlusion or a target deformation happen. In general, part-based trackers are better in addressing these issues. Therefore several part-based correlation filter methods were proposed. \cite{part_cf_cvpr2016} use an efficient method to combine correlation outputs of multiple parts and ~\cite{structural_cf_cvpr2016} proposed a tracking method for modeling the target structure with multiple parts using multiple correlation filters. \cite{lukezic_dpt} treat the parts correlation filter responses and their constellation constraints jointly as an equivalent spring system. They derive a highly efficient optimization to infer the most probable target deformation.

Recently, \cite{cfwlb_cvpr2015} addressed the problem that occurs due to learning with circular correlation from small training regions. They proposed a learning framework that artificially increases the filter size by implicitly zero padding the filter. This reduces the boundary artifacts by increasing the number of training examples in constrained filter learning. \cite{srdcf_iccv2015} reformulate the learning cost function to penalize non-zero filter values outside the object bounding box. Performance better than~\citep{cfwlb_cvpr2015} is reported, but the learned filter is still a trade-off between the correlation response and regularization, and it does not guarantee that filter values are zero outside of object bounding box.

\section{Spatially constrained correlation filters} \label{sec:approach}

The use of multiple channels in correlation filters \citep{henriques2015tracking,danelljan_dsst_pami,galoogahi_multi_channel_correlation} has become very popular in visual tracking. In the following we present the main ideas behind learning these filters. Given a set of $N_c$ channel features $\mathbf{f}=\{ \mathbf{f}_d \}_{d=1:N_c}$ and corresponding target templates (filters) $\mathbf{h}=\{ \mathbf{h}_d \}_{d=1:N_c}$, the object position is estimated as the location of the maximum of correlation response $\tilde{\mathbf{g}}(\mathbf{h})$,
\begin{equation}\label{eq:cf_localization}
	\tilde{\mathbf{g}}(\mathbf{h}) = \sum_{d=1}^{N_c} \mathbf{f}_d \star \mathbf{h}_d.
\end{equation}
The symbol $\star$ represents circular correlation between $\mathbf{f}_d \in \mathcal{R}^{c_w \times c_h}$ and $\mathbf{h}_d\in \mathcal{R}^{c_w \times c_h}$, where $c_w$ and $c_h$ are the training/search region width and height, respectively. The optimal correlation filter $\mathbf{h}$ is estimated by minimizing
\begin{equation}\label{eq:cf_cost_f}
	\mathbf{\varepsilon}(\mathbf{h}) = \| \tilde{\mathbf{g}}(\mathbf{h}) - \mathbf{g} \|^2 + \lambda \| \mathbf{h} \|^2,
\end{equation}
where $\mathbf{g}$ is the desired output $\mathbf{g}\in \mathcal{R}^{c_w \times c_h}$, which is typically a 2-D Gaussian function centered at the target location. Efficient tracking performance is achieved by expressing the cost~(\ref{eq:cf_cost_f}) into the Fourier domain
\begin{equation}\label{eq:cf_cost_f_fourier}
	\mathbf{\varepsilon}(\mathbf{h}) = \| \sum_{d=1}^{N_c} \mathrm{diag}(\hat{\mathbf{f}}_d) \overline{\hat{\mathbf{h}}}_d - \hat{\mathbf{g}} \|^2 + \lambda \sum_{d=1}^{N_c} \| \hat{\mathbf{h}} \|^2,
\end{equation}
where the operator $\hat{\mathbf{a}} = \mathrm{vec}(\mathcal{F}[\mathbf{a}])$ is a Fourier  transform of $\mathbf{a}$ reshaped into a column vector, i.e., $\hat{\mathbf{a}} \in \mathcal{R}^{D \times 1}$, with $D = c_w \cdot c_h$, $\mathrm{diag}( \hat{\mathbf{a}} )$ being a $D \times D$ diagonal matrix formed from $\hat{\mathbf{a}}$ and $\overline{( \cdot )}$ is the complex-conjugate operator. The closed-form solution for $d$-th filter channel $\hat{\mathbf{h}}_d$ which minimizes the cost function~(\ref{eq:cf_cost_f_fourier}) is equal to
\begin{equation}\label{eq:cf_closed_form}
	\hat{\mathbf{h}}_d = \bigg( \mathrm{diag}(\hat{\mathbf{f}}_d) \overline{\hat{\mathbf{g}}} \bigg) \odot^{-1} \bigg( \sum_{d=1}^{N_c} \mathrm{diag}(\hat{\mathbf{f}}_d) \overline{\hat{\mathbf{f}}}_d + \lambda \bigg),
\end{equation}
where $\odot^{-1}$ is element-wise division. The solution~(\ref{eq:cf_closed_form}) considers all feature channels jointly and is used in most of the recent correlation filter trackers. Note that the final response is obtained as summation over correlation responses of all channels (\ref{eq:cf_localization}) and  the location of the maximum in the final response represents the new position of the target.

\begin{figure}[!t]
\centering
\includegraphics[width=\linewidth]{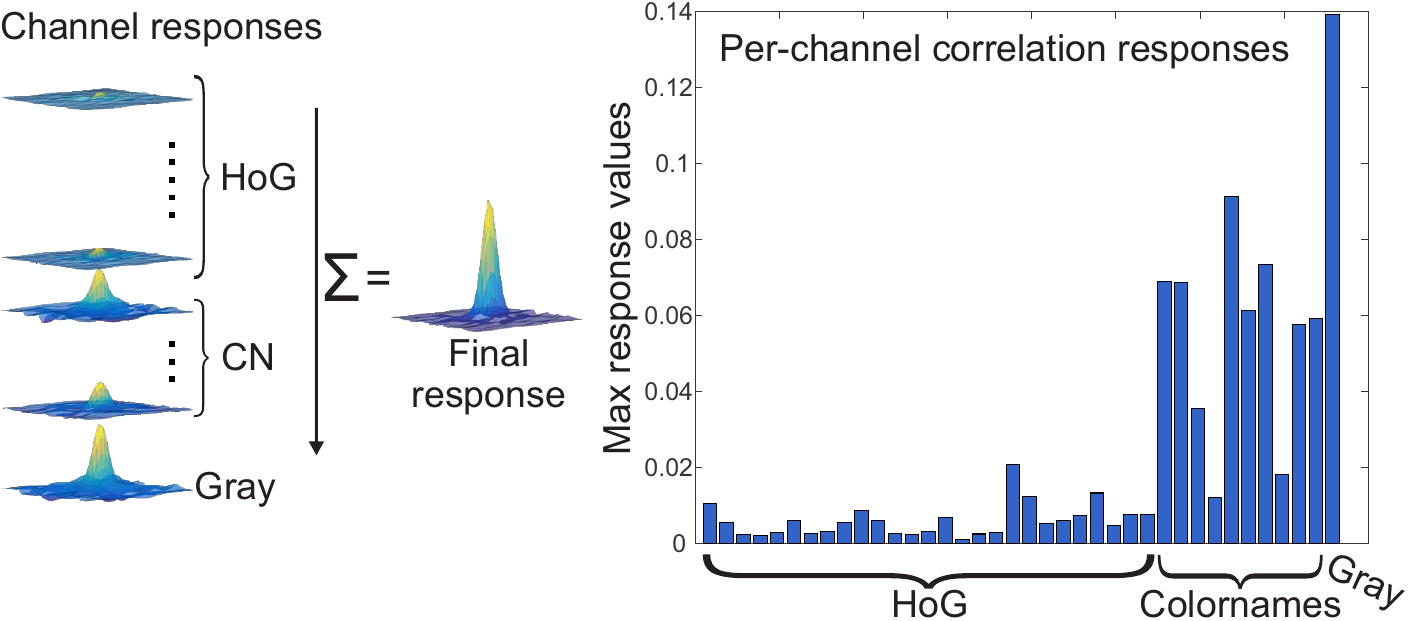}
\caption{Correlation responses of different feature channels (27 HoG, 10 colornames and one grayscale channel) are summed to obtain the final (single channel) correlation response (middle). Note that maximum values of channel responses may vary by orders of magnitude (right).}
\label{fig:max_channels}
\end{figure}

Note that a filter for the $d$-th channel is computed in~(\ref{eq:cf_closed_form})  by dividing $d$-th feature with the sum over all feature channels. This means that the feature {\it scale} crucially impacts the level by which a channel contributes to the final response, irrespective of its discriminative power. Since features (e.g., HoG, colornames and grayscale template) vary in scale, some channels might suppress the others by an order of magnitude. This is demonstrated in Figure~\ref{fig:max_channels} where each HoG channel on its own contributes to the final response very little.


To avoid the issue with different scales we consider each channel independently. This means that each filter channel is optimized to fit the desired output separately. The cost function is thus defined as
\begin{equation}\label{eq:cf_cost_f_indep}
	\mathbf{\varepsilon}(\mathbf{h}) = \sum_{d=1}^{N_c} \| \mathbf{f}_d \star \mathbf{h}_d - \mathbf{g} \|^2 + \lambda \| \mathbf{h}_d \|^2.
\end{equation}
Additionally, we introduce channel weights $\mathbf{w} = \{ \tilde{w}_d \}_{d = 1:N_c}$ which can be considered as scaling factors based on the discriminative power of each feature channel. These weights are called {\it channel reliability weights} in the rest of the paper and they are applied when correlation response is calculated in the target localization stage:
\begin{equation}\label{eq:response_rel}
	\tilde{\mathbf{g}} = \sum_{d=1}^{N_c} \mathbf{f}_d \star \mathbf{h}_d \cdot \tilde{w}_d.
\end{equation}

We present our method for constrained correlation filter learning in Section~\ref{sec:constrained_learning}. The most reliable parts of the filter are identified by introducing the spatial reliability map (Section~\ref{sec:spatial_prior}). The method for channel reliability $w_d$ estimation is presented in Section~\ref{sec:chan_reliab}, the proposed tracker is described in Section~\ref{sec:tracking_framework}.

\subsection{Constrained correlation filter learning} \label{sec:constrained_learning}

Since filter learning is independent across the channels in our formulation (\ref{eq:cf_cost_f_indep}), we assume only a single channel in the following derivation (i.e., $N_c=1$) and drop the channel index for clarity.


Let $\mathbf{m} \in \{0,1\}$ be a spatial reliability map with elements either zero or one, that identifies pixels which should be set to zero in the learned filter. The constraint can be formalized as $\mathbf{h} \equiv \mathbf{m} \odot \mathbf{h}$, where $\odot$ represents the Hadamard (element-wise) product. Such constraint does not lead to a closed-form solution, but an iterative approach akin to~\cite{cfwlb_cvpr2015} can be derived for efficiently solving the optimization problem. In the following we summarize the main steps of our approach and report the full derivation in Appendix~\ref{sec:lagrangian_derivation}.


We start by introducing a dual variable $\mathbf{h}_c$ and the constraint 
\begin{equation}
\mathbf{h}_c - \mathbf{m} \odot \mathbf{h} \equiv 0,
\end{equation}
which leads to the following augmented Lagrangian~\citep{admm_boyd2011}
\begin{eqnarray}\label{eq:augmented_lagrange}
	\mathcal{L}(\hat{\mathbf{h}}_c, \mathbf{h}, \hat{\mathbf{l}} | \mathbf{m}) = \| \mathrm{diag}(\hat{\mathbf{f}}) \overline{\hat{\mathbf{h}}}_c - \hat{\mathbf{g}} \|^2 + \frac{\lambda}{2} \| \mathbf{h}_m \|^2 +\\ \nonumber
[\hat{\mathbf{l}}^H(  \hat{\mathbf{h}}_c - \hat{\mathbf{h}}_m) + \overline{\hat{\mathbf{l}}^H(  \hat{\mathbf{h}}_c -\hat{\mathbf{h}}_m)}] + \mu \|  \hat{\mathbf{h}}_c - \hat{\mathbf{h}}_m \|^2,
\end{eqnarray}
where $\hat{\mathbf{l}}$ is a complex Lagrange multiplier, $\mu > 0$, and we use the definition
$\mathbf{h}_m = (\mathbf{m} \odot \mathbf{h})$ for compact notation. The augmented Lagrangian (\ref{eq:augmented_lagrange}) can be iteratively minimized by the alternating direction method of multipliers, e.g.~\cite{admm_boyd2011}, which sequentially solves the following sub-problems at each iteration: 
\begin{eqnarray}\label{eq:admm_min_hc}
	\hat{\mathbf{h}}_c^{i+1} = \mathop {\arg \min }\limits_\mathbf{h_c} \mathcal{L}	(\hat{\mathbf{h}}_c, \mathbf{h}^{i}, \hat{\mathbf{l}}^{i}| \mathbf{m}),
     \\ \label{eq:admm_min_h}
	\mathbf{h}^{i+1} = \mathop {\arg \min }\limits_\mathbf{h} \mathcal{L}			(\hat{\mathbf{h}}_c^{i+1}, \mathbf{h}, \hat{\mathbf{l}}^{i}| \mathbf{m}),
\end{eqnarray} 
and the Lagrange multiplier is updated as 
\begin{equation}\label{eq:min_l}
\hat{\mathbf{l}}^{i+1} = \hat{\mathbf{l}}^{i} + \mu(\hat{\mathbf{h}}_c^{i+1} - \hat{\mathbf{h}}^{i+1}).
\end{equation}
Minimizations in (\ref{eq:admm_min_hc},\ref{eq:admm_min_h}) have at each iteration a closed-form solution, i.e.,
\begin{equation}\label{eq:min_hc}
\hat{\mathbf{h}}_c^{i+1} = \big( \hat{\mathbf{f}} \odot \overline{\hat{\mathbf{g}}} + (\mu \hat{\mathbf{h}}_m^{i} - \hat{\mathbf{l}}^{i}) \big)  \odot^{-1} \big( \overline{\hat{\mathbf{f}}} \odot \hat{\mathbf{f}} + \mu^{i} \big),
\end{equation}
\begin{equation}\label{eq:min_h}
\mathbf{h}^{i+1} = \mathbf{m} \odot \mathcal{F}^{-1} \big[ \hat{\mathbf{l}}^{i} + \mu^{i} \hat{\mathbf{h}}_c^{i+1} \big] / \big( \frac{\lambda}{2D} + \mu^{i} \big).
\end{equation}
A standard scheme for updating the constraint penalty $\mu$ values~\citep{admm_boyd2011} is applied, i.e., $\mu^{i+1} = \beta \mu^{i}$.

Computations of (\ref{eq:min_hc},\ref{eq:min_l}) are fully carried out in the frequency domain, the solution for (\ref{eq:min_h}) requires a single inverse FFT and another FFT to compute the $\hat{\mathbf{h}}^{i+1}$. A single optimization iteration thus requires only two calls of the Fourier transform, resulting in a very fast optimization. The computational complexity is that of the Fourier transform, i.e., $\mathcal{O}(D\log{}D)$. Filter learning is implemented in less than five lines of Matlab code and is summarized in the Algorithm~\ref{alg:admm_optimization}.

\begin{algorithm}[h!]
\begin{algorithmic}[1]
 \REQUIRE {~}\\
    Features extracted from training region $\mathbf{f}$, ideal correlation response $\mathbf{g}$, \\
    binary mask $\mathbf{m}$.
 \ENSURE {~}\\
    Optimized filter $\mathbf{\widehat{h}}$.
 \\\hspace{-0.6cm}\textbf{Procedure:}
 	\STATE Initialize filter $\mathbf{\widehat{h}}^{0}$ by $\mathbf{h}_{t-1}$. 
 	\STATE Initialize Lagrangian coefficients: $\mathbf{\widehat{l}}^{0}\leftarrow \mathrm{zeros}$.
	\WHILE{stop condition} 
		\STATE{Calculate $\hat{\mathbf{h}}_c^{i+1}$ from $\hat{\mathbf{h}}^{i}$ and $\hat{\mathbf{l}}^{i}$ using ~(\ref{eq:min_hc}).}
		\STATE{Calculate $\mathbf{h}^{i+1}$ from $\hat{\mathbf{h}}_c^{i+1}$ and $\hat{\mathbf{l}}^{i}$ using~(\ref{eq:min_h}).}
		\STATE{Update the Lagrangian $\hat{\mathbf{l}}^{i+1}$ from
        $\hat{\mathbf{h}}_c^{i+1}$ and $\mathbf{h}^{i+1}$~(\ref{eq:min_l}).}
	\ENDWHILE
\end{algorithmic}
\caption{\label{alg:admm_optimization}: Constrained filter optimization.}
\end{algorithm}

\subsection{Constructing spatial reliability map}\label{sec:spatial_prior}

\begin{figure*}[!ht]
\centering
\includegraphics[width=\linewidth]{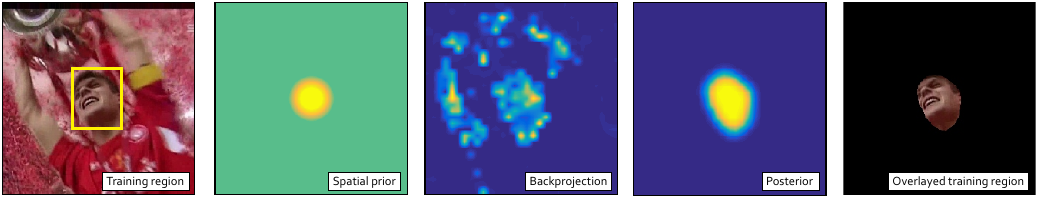}
\caption{Spatial reliability map construction from the training region. From left to right: a training region with the target bounding box, t the foreground-background color models, the posterior object probability after Markov random field regularization, and the training region masked with the final binary reliability map. The probabilities are color-coded in  a blue (0.0) -- green (0.5) -- yellow (1.0) colormap.}
\label{fig:mrfdemo}
\end{figure*}

Once the target is localized, a training region is extracted and used  to update the filter. Our constrained filter learning (\ref{eq:min_h}) requires estimation of spatial reliability map $\mathbf{m}$ (i.e., segmentation) that identifies pixels in the training region which likely belong to the target (see Figure~\ref{fig:mrfdemo}). In the following we briefly outline the segmentation model which is used to estimate $\mathbf{m}$.

During tracking, the object foreground/background color models are maintained as color histograms $\mathbf{c}=\{ \mathbf{c}^\mathrm{f}, \mathbf{c}^\mathrm{b} \}$. Let $\mathbf{y}_i = [\mathbf{y}_{i}^{\mathbf{c}}, \mathbf{y}_{i}^{\mathbf{x}}]$ be the observation, i.e., the color $\mathbf{y}_{i}^{\mathbf{c}}$ and position $\mathbf{y}_{i}^{\mathbf{x}}$ at $i$-th pixel in the training region and let $m_i \in \{0, 1\}$ be a random variable denoting the unknown foreground/background label. The joint probability of observing $y_i$ is defined as 
\begin{eqnarray} \label{eq:mrf1}
	p(\mathbf{y}_i) &=& \sum_{j=0}^{1} p(\mathbf{y}_i | m_i=j) p(m_i=j) = 
	\\
	&=& \sum_{j=0}^{1} p(\mathbf{y}_{i}^{\mathbf{c}} | m_i=j) p(\mathbf{y}_{i}^{\mathbf{x}} | m_i=j) p(m_i=j), \nonumber
\end{eqnarray}
where $p(\mathbf{y}_{i}^{\mathbf{c}} | m_i=j)$, $p(\mathbf{y}_{i}^{\mathbf{x}} | m_i=j)$ and $p(m_i=j)$ are the appearance likelihood, the spatial likelihood and the foreground/background prior probability. The appearance likelihood term $p(\mathbf{y}_{i}^{\mathbf{c}} | m_i=j)$ is computed by Bayes rule from the object foreground/background color models $\mathbf{c}^\mathrm{f}$ and $\mathbf{c}^\mathrm{b}$. The prior probability $p(m_i=j)$ is defined by the ratio between the region sizes for foreground/background histogram extraction.

The central pixels in axis-aligned approximations of an elongated rotating, articulated or deformable object are likely to contain the object regardless of the specific deformation. On the other hand, in the absence of measurements, pixels away from the center belong to the object or background equally likely. This deformation invariance of central elements reliability is enforced in our approach by defining a weak spatial prior 
\begin{equation}\label{eq:inv_prior}
	p(\mathbf{y}_{i}^{\mathbf{x}} | m_i=j) = k(\mathbf{x};\sigma), 
\end{equation}
where $k(\mathbf{x};\sigma)$ is a modified Epanechnikov kernel, $k(r;\sigma)=1-(r/\sigma)^2$, with size parameter $\sigma$ equal to the minor bounding box axis and clipped to interval $[0.5,0.9]$ such that the object prior probability at center is 0.9 and changes to a uniform prior away from the center~(Figure~\ref{fig:mrfdemo}).

\subsubsection{Inference}  \label{sec:mrf_inference}

In practice the likelihood $p(\mathbf{y}_i | m_i)$ is noisy and requires regularization for our filter learning. We thus apply a MRF from \citep{diplaros_genmodel,kristan_tcyb2016}, which treats the prior and posterior label distributions over pixels as random variables and applies a MRF constraint over these. This formulation affords an efficient inference which avoids hard label assignment during optimization and can be implemented as a series of convolutions. 

The prior over the $i$-th pixel is defined compactly as $\pi_i = [\pi_{i0}, \pi_{i1}]$ with $\pi_{ij} = p(m_i = j)$ and a standard approximation is made~\citep{diplaros_genmodel} that decomposes the joint pdf over priors $\pi = [\pi_1,...,\pi_M]$ into a product of local conditional distributions
\(
p(\pi) = \prod\nolimits_{i=1}^{M} p(\pi_i | \pi_{N_i}),
\)
where $M$ is number of pixels, $\pi_{N_i}$ is a mixture distribution over the priors of $i$-th pixel's neighbors, i.e., 
\(
	\pi_{N_i} = \sum\nolimits_{j \in N_i, j\neq i} \lambda_{ij} \pi_j
\)
and $\lambda_{ij}$ are fixed weights satisfying $\sum_{j} \lambda_{ij} = 1$. In \cite{diplaros_genmodel} the weights are fixed to a normalized Gaussian and are shared across all pixel locations. The potentials in the MRF are defined as 
\(
	p(\pi_i | \pi_{N_i}) \propto \exp \big(-\frac{1}{2} E(\pi_i, \pi_{N_i}) \big),
\)
with exponent defined as
\(
	E(\pi_i, \pi_{N_i}) = D(\pi_i || \pi_{N_i}) + H(\pi_i).
\)
The term $D(\pi_i || \pi_{N_i})$ is the Kullback-Leibler divergence which penalizes the difference between prior distributions over the neighboring pixels ($\pi_i$ and $\pi_{N_i}$), while the term $H(\pi_i)$ is the entropy defined as 
\(
	H(\pi_i) = - \sum_{j=0}^{1} \pi_{ij} \log \pi_{ij},
\)
which penalizes uninformative priors $\pi_i$.

For smooth solutions \cite{diplaros_genmodel} propose using a similar constraint over the posteriors $p_i = [p_{i0}, p_{i1}]$ with $p_{ij}$ being the posterior probability of class $j$ at $i$-th pixel, leading to the following energy function 
\begin{equation} \label{eq:mrf6}
	F = \sum_{i=1}^{M} \bigg[ \log p(y_i) -\frac{1}{2}\bigg( E(\pi_i, \pi_{N_i}) + E(p_i, p_{N_i}) \bigg) \bigg].
\end{equation}
Minimization of the energy~(\ref{eq:mrf6}) w.r.t. $\pi$ and $p$ is efficiently solved by the solver from \cite{diplaros_genmodel}. The final mask $\mathbf{m}$ for learning the filter in Section~\ref{sec:constrained_learning} is obtained by thresholding the posterior at 0.5.

\subsection{Channel reliability estimation}\label{sec:chan_reliab}

Channel reliability $\tilde{w}_d$ in~(\ref{eq:response_rel}) reflects the importance of each channel at the target localization stage. In our approach it consists of two types of reliability measures: (i)~{\it channel learning reliability} $\tilde{w}_d^{(\mathrm{lrn})}$, which is calculated in the filter learning stage, and (ii)~{\it channel detection reliability} $\tilde{w}_d^{(\mathrm{det})}$ which is calculated in the target localization stage. The joint channel reliability $\tilde{w}_d$ in~(\ref{eq:response_rel}) at target localization stage is computed as the product of both reliability measures, i.e.,
\begin{equation} \label{eq:channle_rel_all}
	\tilde{w}_d = \tilde{w}_d^{(\mathrm{lrn})} \cdot \tilde{w}_d^{(\mathrm{det})}
\end{equation}
and normalized s.t. $\sum_{d} \tilde{w}_d = 1$. The reliability measures are described in following paragraphs.

\begin{figure}[!ht]
\centering
\includegraphics[width=\linewidth]{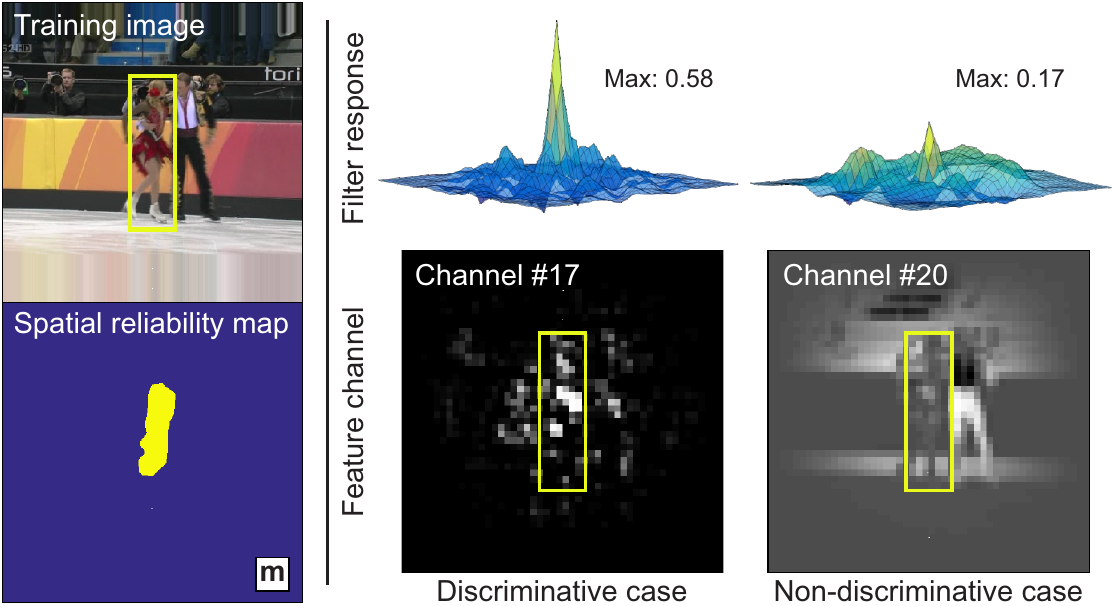}
\caption{A filter is learned on feature channels from a training region using the constrained optimization with a binary segmentation mask $\mathbf{m}$. Correlation responses between the learned filter and the training region for two feature channels are shown on the right. On a discriminative feature channel the filter response is much stronger and less noisy than on a non-discriminative channel.}
\label{fig:channel_rel}
\end{figure}

{\bf Channel learning reliability}.
Constrained minimization of (\ref{eq:augmented_lagrange}) solves a least squares problem averaged over all circular displacements of the filter on a feature channel. A discriminative feature channel $\mathbf{f}_d$ produces a filter $\mathbf{h}_d$ whose output $\mathbf{f}_d*\mathbf{h}_d$ nearly exactly fits the ideal response $\mathbf{g}$. On the other hand, since the response is highly noisy on channels with low discriminative power, a global error reduction in the least squares significantly reduces the maximal response. This effect is demonstrated in Figure~\ref{fig:channel_rel}, which shows correlation responses for a highly discriminative and non-discriminative channels. Thus a straight-forward measure of channel learning reliability $\tilde{w}_d^{(\mathrm{lrn})}$ is the maximum response value of a learned channel filter, which is computed as
\begin{equation} \label{eq:learning_reliability_w}
	\tilde{w}_d^{(\mathrm{lrn})} = \max(\mathbf{f}_d*\mathbf{h}_d).
\end{equation}

{\bf Channel detection reliability}.
The second part of the channel reliability reflects how uniquely each channel votes for a single target location. Note that \cite{bolme2010visual} proposed a similar approach to detect loss of  target. Our measure is based on the ratio between the second and first highest non-adjacent peaks in the channel response map, i.e., $1 - \rho_{d}^{\mathrm{max2}} / \rho_{d}^{\mathrm{max1}}$. The two largest peaks in the response map are obtained as two largest values after a $3\times 3$ non-maximum suppression. Note that this ratio penalizes situations in which multiple similar objects appear in the target vicinity (i.e., response map contains many well expressed modes), even if the major mode accurately depicts the target position. To mitigate such penalizations, the final values are note allowed to fall below $0.5$. The detection reliability of $d$-th channel is estimated as
\begin{equation} \label{eq:detection_reliability_w}
	\tilde{w}_d^{(\mathrm{det})}= \max(1 - \rho_{d}^{\mathrm{max2}} / \rho_{d}^{\mathrm{max1}}, 0.5).
\end{equation}

\subsection{Tracking with channel and spatial reliability}  \label{sec:tracking_framework}

\begin{figure*}[!t]
\centering
\includegraphics[width=\linewidth]{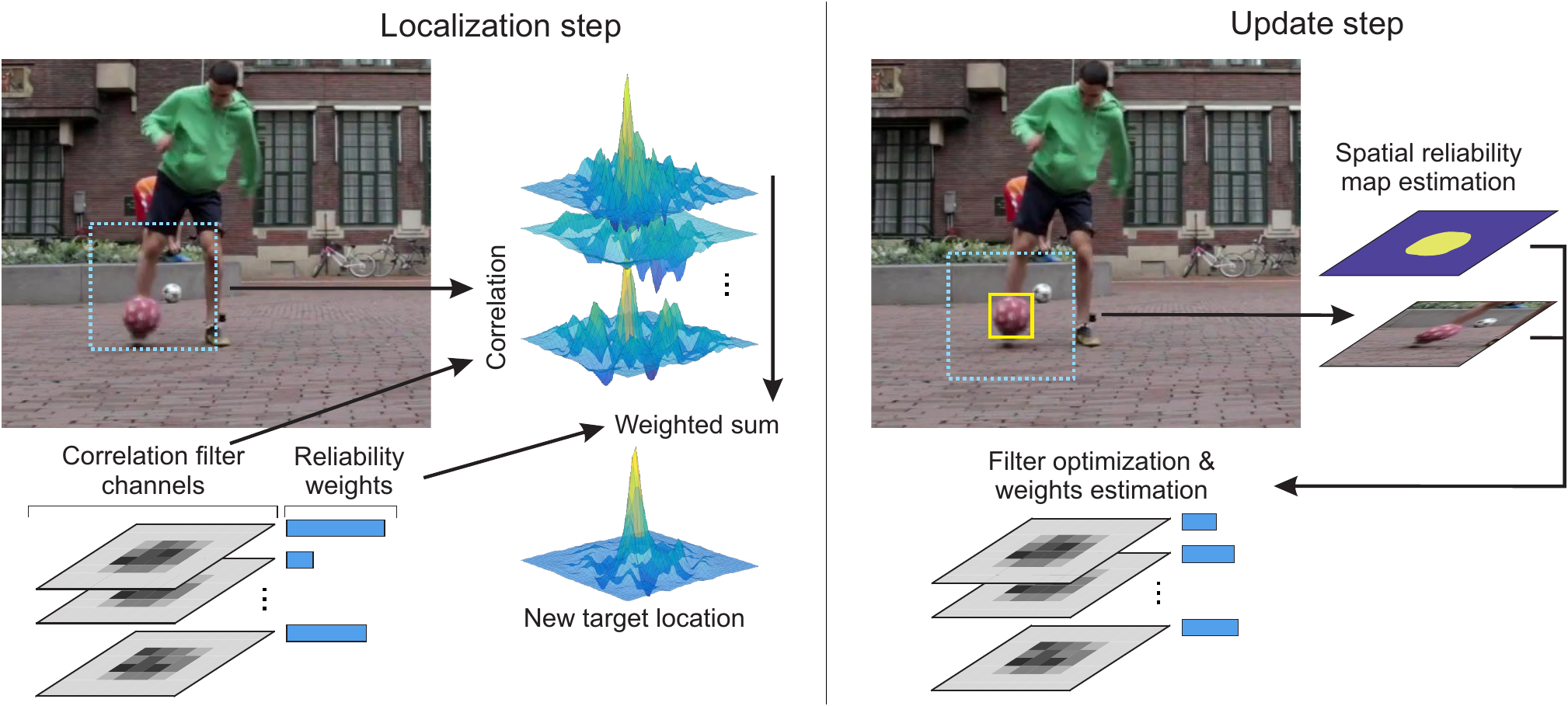}
\caption{The CSR-DCF tracking iteration: localization step is shown on the left and update step on the right side of the image.}
\label{fig:csr_scheme}
\end{figure*}
A single tracking iteration of the proposed channel and spatial reliability correlation filter tracker (CSR-DCF) is summarized in Algorithm~\ref{alg:tracking_algorithm} and visualized in Figure~\ref{fig:csr_scheme}. The localization and update steps proceed as follows.

\textbf{Localization step.} Features are extracted from a search region centered at the target estimated position in the previous time-step and correlated with the learned filter $\mathbf{h}_{t-1}$. The object is localized by summing the correlation responses weighted by the estimated channel reliability scores $\mathbf{w}_{t-1}$. The scale is estimated by a single scale-space correlation filter as in \cite{danelljan2014accurate}. Per-channel filter responses are used to compute the corresponding detection reliability values $\tilde{\mathbf{w}}^{(\mathrm{det})}=[\tilde{w}^{(\mathrm{det})}_1, \dots, \tilde{w}^{(\mathrm{det})}_{N_c}]^T$ according to~(\ref{eq:detection_reliability_w}).

\textbf{Update step.} The training region is centered at the target location estimated at localization step. The foreground and background histograms $\tilde{\mathbf{c}}$ are extracted and updated by exponential moving average with learning rate $\eta_c$ (step \ref{alg-step:update_histograms} in Algorithm~\ref{alg:tracking_algorithm}). The foreground histogram is extracted by an Epanechnikov kernel within the estimated object bounding box and the background is extracted from the neighborhood twice the object size. The spatial reliability map $\mathbf{m}$ (Sect.~\ref{sec:spatial_prior}) is constructed and the optimal filters $\tilde{\mathbf{h}}$ are computed by optimizing (\ref{eq:augmented_lagrange}). The per-channel learning reliability weights $\tilde{\mathbf{w}}^{(\mathrm{lrn})}=[\tilde{w}^{(\mathrm{lrn})}_1, \dots, \tilde{w}^{(\mathrm{lrn})}_{N_c}]^T$ are estimated from the correlation responses (\ref{eq:learning_reliability_w}). Current frame reliability weights $\tilde{\mathbf{w}}$ are computed from detection and learning reliability (\ref{eq:channle_rel_all}). The filters and channel reliability weights are updated by exponential moving average (current and from previous frame) with learning rate $\eta$ (steps \ref{alg-step:update_h} and \ref{alg-step:update_w} in the Algorithm~\ref{alg:tracking_algorithm}). Note that we compute the spatial reliability map in each frame independently to capture large target appearance changes, e.g. caused by rotation or deformation.

\begin{algorithm}[!t]
\begin{algorithmic}[1]
 \REQUIRE {~}\\
    Image $\mathbf{I}_{t}$, object position on previous frame $\mathbf{p}_{t-1}$, scale $s_{t-1}$, filter $\mathbf{h}_{t-1}$, color histograms $\mathbf{c}_{t-1}$, channel reliability $\mathbf{w}_{t-1}$.
 \ENSURE {~}\\
    Position $\mathbf{p}_{t}$, scale $s_{t}$ and updated models.
 \\\hspace{-0.6cm}\textbf{Localization and scale estimation:}
 	\STATE New target location $\mathbf{p}_{t}$: position of the maximum in correlation between $\mathbf{h}_{t-1}$ and image patch features $\mathbf{f}$ extracted on position $\mathbf{p}_{t-1}$
 and weighted by the channel reliability scores $\mathbf{w}$ (Sect.~\ref{sec:chan_reliab}).    
    \STATE Using per-channel responses, estimate detection reliability $\tilde{\mathbf{w}}^{(\mathrm{det})}$ (Sect.~\ref{sec:chan_reliab}).
	\STATE Using location $\mathbf{p}_{t}$, estimate new scale $s_{t}$.
 \\\hspace{-0.6cm}\textbf{Update:}
 	\STATE Extract foreground and background histograms $\tilde{\mathbf{c}}^{f}$, $\tilde{\mathbf{c}}^{b}$.
  	\STATE Update foreground and background histograms \\ $\mathbf{c}_{t}^{f} = (1-\eta_c)\mathbf{c}_{t-1}^{f} + \eta_c\tilde{\mathbf{c}}^{f}$, $\mathbf{c}_{t}^{b} = (1-\eta_c)\mathbf{c}_{t-1}^{b} + \eta_c\tilde{\mathbf{c}}^{b}$.\label{alg-step:update_histograms}
	\STATE Estimate reliability map $\mathbf{m}$  (Sect.~\ref{sec:spatial_prior}).
	\STATE Estimate a new filter $\tilde{\mathbf{h}}$ using $\mathbf{m}$ (Algorithm~\ref{alg:admm_optimization}).
	\STATE Estimate learning channel reliability $\tilde{\mathbf{w}}^{(\mathrm{lrn})}$ from $\mathbf{h}$ (Sect.~\ref{sec:chan_reliab}).
    \STATE Calculate channel reliability $\tilde{\mathbf{w}} = \tilde{\mathbf{w}}^{(\mathrm{lrn})} \odot \tilde{\mathbf{w}}^{(\mathrm{det})}$
    \STATE Update filter $\mathbf{h}_{t} = (1-\eta)\mathbf{h}_{t-1} + \eta\tilde{\mathbf{h}}$.\label{alg-step:update_h}
    \STATE Update channel reliability $\mathbf{w}_{t} = (1-\eta)\mathbf{w}_{t-1} + \eta\tilde{\mathbf{w}}$.\label{alg-step:update_w}
\end{algorithmic}
\caption{\label{alg:tracking_algorithm}: The CSR-DCF tracking algorithm.}
\end{algorithm}

\subsection{Comparison with prior work}  \label{sec:differences}

\cite{cfwlb_cvpr2015} and \cite{srdcf_iccv2015} have previously considered constrained filter learning. Here we highlight the differences of our approach.

The LBCF tracker~\citep{cfwlb_cvpr2015} addresses the circular boundary effect of the Fourier transform and implicitly increases the filter search region size. In contrast, the CSR-DCF primarily reduces the impact of the background in the filter. The solution of~\cite{cfwlb_cvpr2015} is similar to our filter optimization, but it is derived for a rectangular mask only. Since rotating and deformable targets are poorly approximated by an axis-aligned bounding box their filter is contaminated by background leading to a reduced performance. The LBCF updates the auto-spectral and cross-spectral energies ($\hat{\mathbf{f}} \odot \overline{\hat{\mathbf{f}}}$ and $\hat{\mathbf{f}} \odot \overline{\hat{\mathbf{g}}}$ in~(\ref{eq:min_hc})) separately, which approximates computation of a single filter from a weighted sum of errors over past training samples. This adaptation is reasonable since it is derived for a rectangular mask that remains constant throughout tracking. The CSR-DCF estimates the mask separately for each training sample and learns a corresponding filter. For articulated objects in particular the mask varies significantly with time, therefore it is beneficial to compute the exact filter for each frame. Robustness is increased by moderately averaging the filters temporally.

Similarly to our approach, the SRDCF \citep{srdcf_iccv2015} uses a spatial map in filter learning. In contrast to our approach, their map does not adapt to the target and is required to be highly smooth for their optimization to converge. In CSR-DCF the map serves as a hard constraint resulting in a filter with values off the target set to zero. In contrast, the SRDCF \citep{srdcf_iccv2015} filter is a compromise between target position regression and a penalty term that {\it prefers} potentially non-zero values in the filter center and close-to-zero values away from the center, but does not guarantee zero values outside the mask.

\section{Experimental analysis}  \label{sec:experimental_analysis}

This section presents a comprehensive experimental evaluation of the  CSR-DCF tracker. Implementation details are discussed in Section~\ref{sec:implementation_details}, convergence of the filter optimization method is presented in Section~\ref{sec:convergence_exp}, Section~\ref{sec:boundary_impact} reports comparison of the proposed constrained learning to the related state-of-the-art and the ablation study is provided in Section~\ref{sec:ablation}. Tracking performance on three recent benchmarks: OTB-100~\citep{otb_pami2015}, VOT2015~\citep{kristan_vot2015} and VOT2016~\citep{kristan_vot2016} is reported in Sections~\ref{sec:otb_2015}, \ref{sec:vot_2015} and~\ref{sec:vot_2016}, respectively. The detailed analysis of the tracker, including per-attribute tracking performance is presented in Section~\ref{sec:vot_2016_per_attribute} and tracking speed analysis in Section~\ref{sec:realtime_2016}.
 
\subsection{Implementation details and parameters} \label{sec:implementation_details}

A popular implementation~\cite{felzenszwalb_dpm} of the standard HoG~\citep{dalal_triggs_hog} and Colornames \citep{colornames_tip2009} features are used in the correlation filter and HSV foreground/background color histograms with $16$ bins per color channel are used in reliability map estimation with parameter $\alpha_\mathrm{min}=0.05$. All the parameters are set to values commonly used in literature~\citep{srdcf_iccv2015,cfwlb_cvpr2015}. Histogram adaptation rate is set to $\eta_c=0.04$, correlation filter adaptation rate is set to $\eta = 0.02$, and the regularization parameter is set to $\lambda=0.01$. The augmented Lagrangian optimization parameters are set to $\mu^{0}=5$ and $\beta=3$. All parameters have a straight-forward interpretation, do not require fine-tuning, and were kept constant throughout all experiments.
Our Matlab implementation\footnote{The CSR-DCF Matlab source is publicly available on: \\ \url{https://github.com/alanlukezic/csr-dcf}} runs at $13$ frames per second on an Intel Core i7 3.4GHz standard desktop.

\subsection{Convergence of constrained learning} \label{sec:convergence_exp}

The constrained filter learning described in Section~\ref{sec:constrained_learning} is an iterative optimization method that minimizes the cost function~(\ref{eq:augmented_lagrange}). This experiment demonstrates how the cost changes with the number of iterations during filter optimization.

\begin{figure}[!ht]
\centering
\includegraphics[width=.75\linewidth]{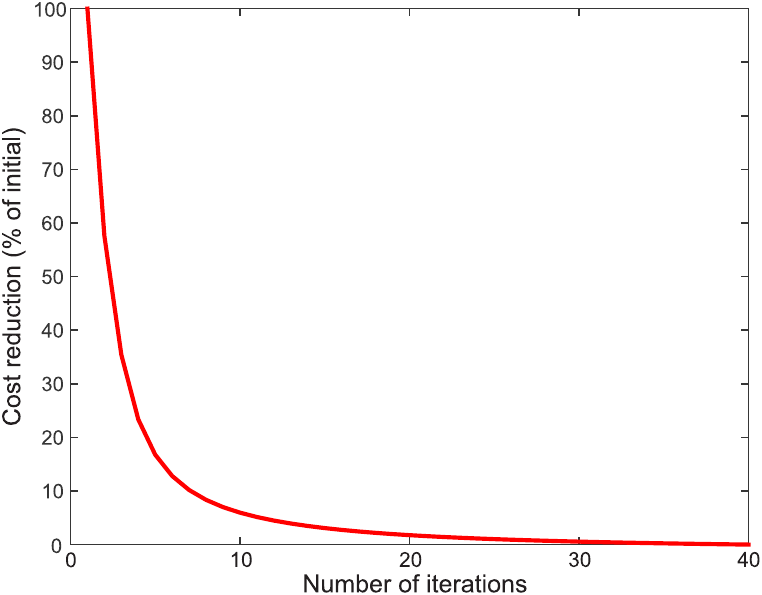}
\caption{Convergence speed of constrained filter learning from Section~\ref{sec:constrained_learning} shown as a relative drop of the initial cost.}
\label{fig:convergence}
\end{figure}

Figure~\ref{fig:convergence} shows the average squared difference between the result of the correlation of the filter constrained by the spatially constrained function and the ideal output. This graph was obtained by averaging 60 examples of initializing a filter on a target (one per VOT2015 sequence) and scaling each to an interval between zero and one. It is clear that the error drops by $80\%$ within the first few iterations. Already after four iterations, the performance improvements become negligible, therefore we set number of iterations to $N=4$.

\subsection{Impact of the boundary constraint formulation}  \label{sec:boundary_impact}

This section compares our proposed boundary constraints formulation (Sect.~\ref{sec:approach}) with recent state-of-the-art approaches \citep{srdcf_iccv2015,cfwlb_cvpr2015}. In the first experiment, three variants of the standard single-scale HoG-based correlation filter were implemented to emphasize the difference in boundary constraints: the first uses our spatial reliability boundary constraint formulation from Section~\ref{sec:approach} ($\mathrm{T}_\mathrm{SC}$) the second applies the spatial regularization constraint~\citep{srdcf_iccv2015} ($\mathrm{T}_\mathrm{SR}$) and the third applies the limited boundaries constraint~\citep{cfwlb_cvpr2015} ($\mathrm{T}_\mathrm{LB}$).

The three variants were compared on the challenging VOT2015 dataset~\citep{kristan_vot2015} by applying a standard no-reset one-pass evaluation from OTB~\citep{otb_cvpr2010} and computing the AUC on the success plot. The tracker with our constraint formulation $\mathrm{T}_\mathrm{SC}$ achieved 0.32 AUC, while the alternatives achieved 0.28 ($\mathrm{T}_\mathrm{SR}$) and 0.16 ($\mathrm{T}_\mathrm{LB}$). The only difference between these tackers is in the constraint formulation, which indicates superiority of the proposed spatial-reliability-based constraints formulation over the recent alternatives~\citep{cfwlb_cvpr2015,srdcf_iccv2015}.

\subsubsection{Robustness to non-axis-aligned target initialization}

The CSR-DCF tracker from Section~\ref{sec:approach} was compared to the original recent state-of-the-art trackers SRDCF \citep{srdcf_iccv2015} and LBCF~\citep{cfwlb_cvpr2015} that apply alternative boundary constraints. For fair comparison, the source code of SRDCF and LBCF was obtained from the authors, all three trackers used only HoG features and tracked on the same single scale. An experiment was designed to evaluate initialization and tracking of non axis-aligned targets, which is the case for most realistic deforming and non-circular objects. Trackers were initialized on frames with non-axis aligned targets and left to track until the sequence end, resulting in a large number of tracking trajectories.

\begin{figure}[!t]
\centering
\includegraphics[width=\linewidth]{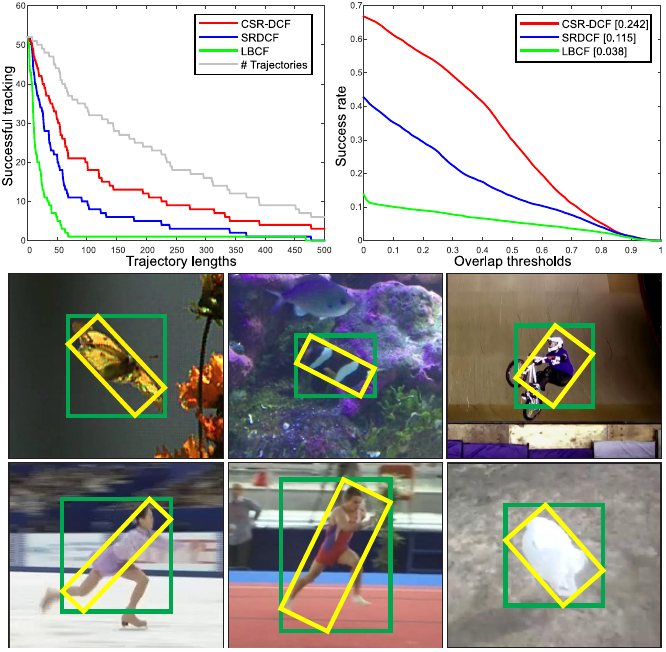}
\caption{The number of trajectories with tracking successful up to frame $\Theta_\mathrm{frm}$ (upper left), the success plots (upper right) and initialization examples of non-axis-aligned targets (bottom).}
\label{fig:init_experiment}
\end{figure}

The VOT2015 dataset~\citep{kristan_vot2015} contains non-axis-aligned annotations, which allows automatic identification of tracker initialization frames, i.e., frames in which the ground truth bounding box significantly deviates from an axis-aligned approximation. Frames with overlap (intersection over union of predicted and ground-truth bounding boxes) of the ground truth and the axis-aligned approximation lower than $0.5$ were identified and filtered to obtain a set of initialization frames at least hundred frames apart. This constraint fits half the typical short-term sequence length~\citep{kristan_vot2015} and reduces the potential correlation across the initializations (see Figure~\ref{fig:init_experiment} (bottom) for examples).

\begin{table}[!t]
\begin{center}
\caption{Comparison of three most related trackers on non-axis-aligned initialization experiment: weighted average tracking length in frames $\Gamma_\mathrm{frm}$ and proportions $\Gamma_\mathrm{prp}$, and weighted average overlaps using the original and axis-aligned ground truth, $\Phi_\mathrm{rot}$ and $\Phi_\mathrm{aa}$, respectively.}
\label{tab:initi_experiment}
\begin{tabular}{l r r r r}
\hline
Tracker & \multicolumn{1}{c}{$\Gamma_\mathrm{prp}$} & \multicolumn{1}{c}{$\Gamma_\mathrm{frm}$} & \multicolumn{1}{c}{$\Phi_{\mathrm{aa}}$} & \multicolumn{1}{c}{$\Phi_{\mathrm{rot}}$} \\
\hline
CSR-DCF & \first{0.58} & \first{221} & \first{0.31} & \first{0.24} \\
SRDCF \tiny{(ICCV2015)} & \second{0.31} & \second{~~95} & \second{0.16} & \second{0.12} \\
LBCF \tiny{(CVPR2015)}& \third{0.12} & \third{~~37} & \third{0.06} & \third{0.04} \\
\hline
\end{tabular}
\end{center}
\end{table}

\begin{figure}[th]
\centering
\includegraphics[width=\linewidth]{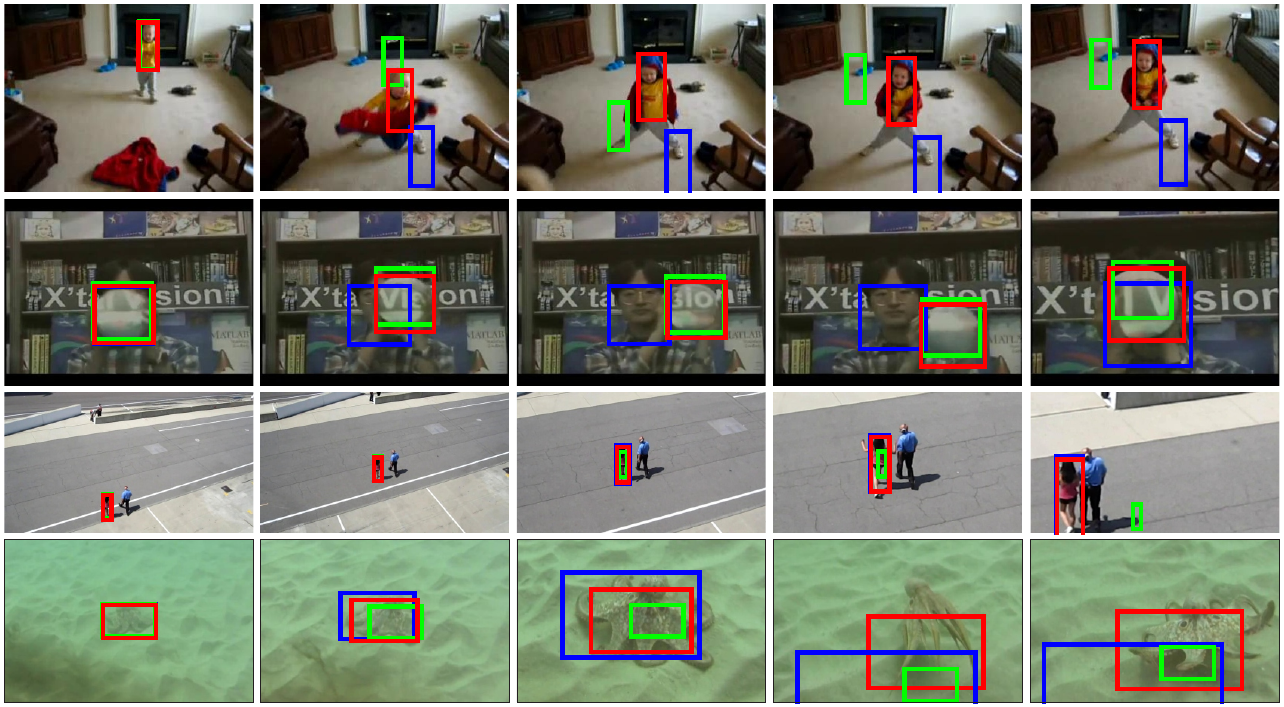}
\caption{Qualitative results for trackers CSR-DCF (red) tracker, SRDCF (blue) and LBCF (green).}
\label{fig:thumbs}
\end{figure}

Initialization robustness is estimated by counting the number of trajectories in which the tracker was still tracking (overlap with ground truth greater than 0) $\Theta_\mathrm{frm}$ frames after initialization. The graph in Figure~\ref{fig:init_experiment} (top-left) shows these values with increasing the threshold $\Theta_\mathrm{frm}$. The CSR-DCF graph is consistently above the SRDCF and LBCF for all thresholds. The performance is summarized by the average tracking length (number of frames before the overlap drops to zero) weighted by trajectory lengths. The weighted average tracking lengths in frames, $\Gamma_\mathrm{frm}$, and proportions of full trajectory lengths, $\Gamma_\mathrm{prp}$, are shown in Table~\ref{tab:initi_experiment}. The CSR-DCF by far outperforms SRDCF and LBCF in all measures indicating significant robustness in the initialization of challenging targets that deviate from axis-aligned templates. 
This improvement is further confirmed by the graph in Figure~\ref{fig:init_experiment} (top-right) which shows the OTB success plots \citep{otb_cvpr2010} calculated on these trajectories and summarized by the AUC values, which are equal to the average overlaps \citep{cehovin_tip2016}. Table~\ref{tab:initi_experiment} shows the average overlaps computed on the original ground truth on VOT2015 ($\Phi_\mathrm{rot}$) and on ground truth approximated by the axis-aligned bounding box ($\Phi_\mathrm{aa}$). Again, the CSR-DCF by far outperforms the competing alternatives SRDCF and LBCF. Tracking examples for the three trackers are shown in Figure~\ref{fig:thumbs}.

In summary, the results show that the quality of spatial constraints significantly affects the relative tracking performance when a large portion of the training region in the target vicinity is occupied by background. The relative performance of LBCF~\citep{cfwlb_cvpr2015} is lowest among the three trackers since this tracker treats all pixels within axis-aligned bounding box equally as target. The SRDCF~\citep{srdcf_iccv2015} mostly focuses on the central pixels of the training region and suppresses the filter values at the borders, thus outperforming the LBCF~\citep{cfwlb_cvpr2015}. The spatial reliability map in CSR-DCF most successfully reduces the influence of background in filter learning resulting in considerable robustness to poor initializations.

\subsection{Spatial and channel reliability ablation study} \label{sec:ablation}

An ablation study on VOT2016 was conducted to evaluate the contribution of spatial and channel reliability in CSR-DCF. Results of the VOT primary measure expected average overlap (EAO) and two supplementary measures accuracy and robustness (A,R) are summarized in Table~\ref{tab:tracker_analysis}. For the details of  performance measures and evaluation protocol we refer the reader to the Section~\ref{sec:vot_2015}. Performance of the various modifications of  CSR-DCF is discussed in the following.

\noindent {\bf Channel reliability weights}.
Setting the channel reliability weights to uniform values (CSR-DCF$\mathrm{_{c^{-}}}$) is equivalent to treating all channels as independent and equally important. The performance drop in EAO compared to CSR-DCF is $12\%$.

\noindent {\bf Spatial reliability map}.
Replacing the spatial reliability map in CSR-CDF by a constant map with uniform values within the bounding box and zeros elsewhere (CSR-DCF$\mathrm{_{s^{u}}}$), results in a $21\%$ drop in EAO. The other parts of the tracker remained unchanged in this experiment, including the channel reliability. This clearly shows the importance of our segmentation-based spatial reliability map estimation from Section~\ref{sec:spatial_prior}.

\noindent {\bf Channel and spatial reliability}.
Making both replacements in the original tracker means that this version (CSR-DCF$\mathrm{_{c^{-}s^{u}}}$) does not use channel reliability weights and it uses uniform spatial reliability map (uniform values within the bounding box and zeros elsewhere). The performance drops by $24\%$ compared to CSR-DCF. Removal of the uniform spatial reliability map from CSR-DCF$\mathrm{_{c^{-}s^{u}}}$ results in the CSR-DCF$\mathrm{_{c^{-}s^{-}}}$. This version reduces our tracker to a standard DCF with a large receptive field. Since the learned filter captures a significant amount of background, the  performance drops by over $50\%$.

\noindent {\bf ADMM Filter optimization method}.
To demonstrate the importance of the constrained optimization method we modify the proposed tracker as follows. The filter $\mathbf{h}$ is calculated with a naive approach, i.e., a closed-form solution followed by masking with the spatial reliability map $\mathbf{m}$: $\hat{\mathbf{h}} = \mathcal{F}(\mathcal{F}^{-1}(\hat{\mathbf{h}}) \odot \mathbf{m})$. For a fair comparison the tracker, denoted as CSR-DCF$\mathrm{_{c^{-}o^{-}}}$, does not use channel reliability weights. The performance drop in EAO compared to CSR-DCF$\mathrm{_{c^{-}}}$ is $15\%$.

\begin{table}\setlength{\tabcolsep}{1pt}
\begin{center}
\caption{Ablation study of CSR-DCF. The use of channel reliability is indicated in the  {\it Chan.} column, the the type of spatial reliability map  in the {\it Spat.}  column. The {\it Opt.} column indicates whether the constrained optimization is used.}
\label{tab:tracker_analysis}
\begin{tabular*}{1\linewidth}{l c c c @{\hspace{3\tabcolsep}} r @{\hspace{3\tabcolsep}} r @{\hspace{3\tabcolsep}} r}
\hline
 Tracker & Chan. & Spat. & Opt. & \multicolumn{1}{c}{EAO} & \multicolumn{1}{c}{$R_\mathrm{av}$} & \multicolumn{1}{c}{$A_\mathrm{av}$} \\
\hline
CSR-DCF & \cmark & segm. & \cmark   & \first{0.338} & \first{0.85} & \first{0.51} \\
CSR-DCF$\mathrm{_{c^{-}}}$ & \xmark & segm. & \cmark  & \second{0.297} & \second{1.08} & \second{0.51} \\
CSR-DCF$\mathrm{_{s^{u}}}$ & \cmark & unif. & \cmark & \third{0.264} & \third{1.18} & \third{0.49} \\
CSR-DCF$\mathrm{_{c^{-}s^{u}}}$ & \xmark & unif. & \cmark &0.256 & 1.33 & \second{0.51} \\
CSR-DCF$\mathrm{_{c^{-}o^{-}}}$ & \xmark & segm. & \xmark & 0.251 & 1.47 & \second{0.51} \\
CSR-DCF$\mathrm{_{c^{-}s^{-}}}$ & \xmark & \xmark & \xmark & 0.152 & 2.85 & 0.47 \\
\hline
\end{tabular*}
\end{center}
\end{table}


\subsection{Spatial reliability map quality analysis}

In this section we evaluate the quality of our spatial reliability map estimation (Section~\ref{sec:spatial_prior}) from a visual tracking perspective. We compare the CSR-DCF tracker with the version of CSR-DCF that uses {\it ideal} spatial reliability map (the tracker is denoted as CSR*-DCF). In the VOT2016 challenge~\citep{kristan_vot2016}, the ground truth bounding boxes were automatically computed by optimizing coverage over manually segmented targets in each frame. The VOT2016 has recently made their per-frame segmentations freely available~\citep{vojir_segmentation}. We use these per-frame segmentation masks in CSR*-DCF as spatial reliability map $\mathbf{m}$. 

Results of evaluation on VOT2016~\citep{kristan_vot2016} are reported in Table~\ref{tab:segmentation-quality}. The performances of the CSR-DCF and CSR*-DCF are very similar. The trackers achieve an equal expected average overlap (EAO) and average accuracy ($A_\mathrm{av}$). But the CSR*-DCF has a single failure less than CSR-DCF on 60 sequences which is 0.02 on average. In Table~\ref{tab:segmentation-quality} the average number of failures is denoted as robustness ($R_\mathrm{av}$). These results show that our approach for spatial reliability estimation (Section~\ref{sec:spatial_prior}) generates near ideal maps from a tracking perspective. 

\begin{table}[t!]\setlength{\tabcolsep}{10pt}
\begin{center}
\caption{Tracking performance comparison of the two versions of CSR-DCF on VOT2016. The proposed method is denoted as CSR-DCF while the version using ground-truth segmentation masks instead of color-based spatial reliability map is denoted as CSR*-DCF.}
\label{tab:segmentation-quality}
\begin{tabular*}{0.8\linewidth}{l c c c}
\hline
 Tracker & EAO & $A_\mathrm{av}$ & $R_\mathrm{av}$ \\
\hline
CSR-DCF & 0.338 & 0.51 & 0.85 \\
CSR*-DCF & 0.338 & 0.51 & 0.83 \\
\hline
\end{tabular*}
\end{center}
\end{table}

\begin{figure}[!t]
\centering   
\includegraphics[width=\linewidth]{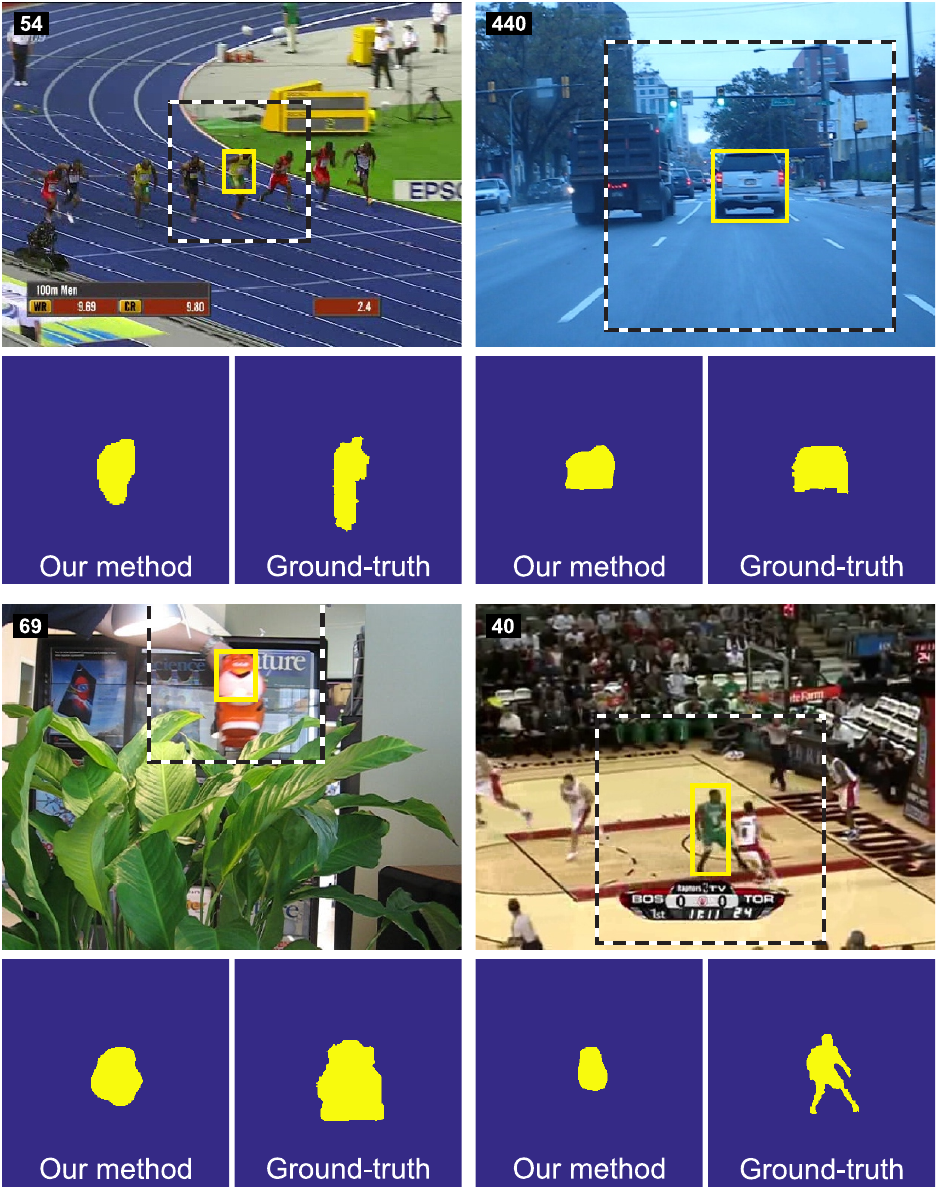}
\caption{Qualitative comparison of the spatial reliability maps during tracking. The dashed bounding box represents area from which correlation filter is obtained. This is also the area where spatial reliability map is calculated. In addition, the ground-truth segmentation masks are visualized on the right side under each frame.}
\label{fig:segmentation-quality}
\end{figure}

Figure~\ref{fig:segmentation-quality} qualitatively compares the spatial reliability maps to the ground-truth segmentation masks on VOT2016 \citep{kristan_vot2016}. Note that at pixel level, the maps are different. But from the perspective of tracking they are nearly equivalent since the tracking performance remains unchanged. For example, in the case of a basketball player, the legs are not well segmented by our approach. But since the legs constantly move, they are in fact non-informative for object localization from the perspective of the correlation filter template matching and do not contribute to improved tracking.

\subsection{The OTB100 benchmark~\citep{otb_pami2015}} \label{sec:otb_2015}

The OTB100~\citep{otb_pami2015} benchmark contains results of $29$ trackers evaluated on $100$ sequences by a no-reset evaluation protocol. Tracking quality is measured by precision and success plots. The success plot shows the fraction of frames with the overlap between th epredicted and ground truth bounding box greater than a threshold with respect to all threshold values. The precision plot shows similar statistics on the center error. The results are summarized by areas under these plots. To reduce clutter, we show here only the results for top-performing recent baselines, i.e., Struck~\citep{hare_struck}, TLD~\citep{kalal_pami}, CXT~\citep{cxt_cvpr2011}, ASLA~\citep{asla_cvpr2012}, SCM~\citep{scm_cvpr2012}, LSK~\citep{lsk_cvpr2011}, CSK~\citep{csk_henriques_eccv2012} and results for recent top-performing state-of-the-art trackers SRDCF~\citep{srdcf_iccv2015} and MUSTER~\citep{muster_cvpr2015}.

\begin{figure}
\centering   
\includegraphics[width=\linewidth]{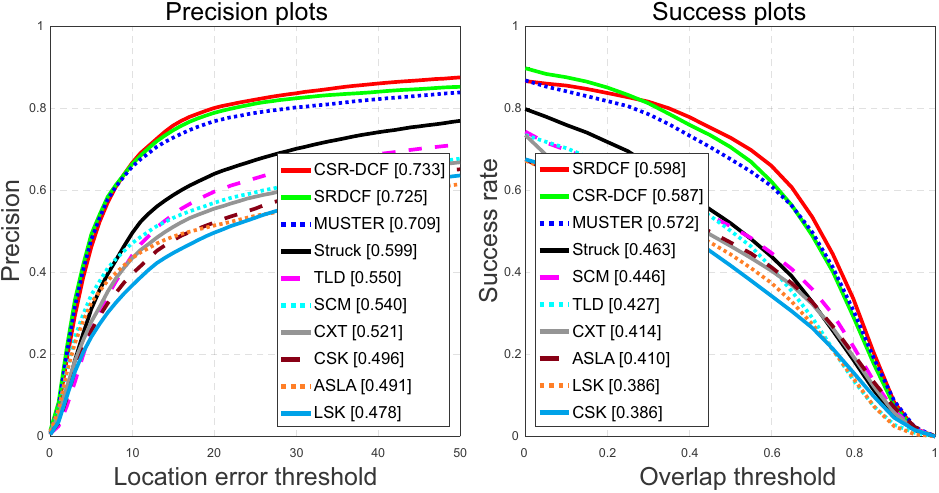}
\caption{Evaluation on OTB100~\citep{otb_pami2015} benchmark.}
\label{fig:otb_graphs}
\end{figure}

The CSR-DCF is ranked top on the benchmark (Fig.~\ref{fig:otb_graphs}). It significantly outperforms the best performers reported in \citep{otb_pami2015} and outperforms the current state-of-the-art SRDCF~\citep{srdcf_iccv2015} and MUSTER \citep{muster_cvpr2015}. The average CSR-DCF performance on success plot is slightly lower than SRDCF~\citep{srdcf_iccv2015} due to poorer scale estimation, but yields better performance in the average precision (center error). Both, precision and success plot, show that the CSR-DCF tracks on average longer than competing methods.

\subsection{The VOT2015 benchmark~\citep{kristan_vot2015}} \label{sec:vot_2015} 

The VOT2015~\citep{kristan_vot2015} benchmark contains results of 
63 state-of-the-art trackers evaluated on 60 challenging sequences.
In contrast to related benchmarks, the VOT2015 dataset was constructed from over 300 sequences by an advanced sequence selection methodology that favors objects difficult to track and maximizes a visual attribute diversity cost function~\citep{kristan_vot2015}. This makes it arguably the most challenging sequence set available.  The VOT methodology~\citep{kristan_vot_tpami2016} resets a tracker upon failure to fully use the dataset. The basic VOT measures are the number of failures during tracking (robustness) and average overlap during the periods of successful tracking (accuracy), while the primary VOT2015 measure is the expected average overlap (EAO) on short-term sequences. The latter can be thought of as the expected no-reset average overlap (AUC in OTB methodology), but with reduced bias and the variance as explained in~\citep{kristan_vot2015}.

\begin{figure}[!t]
	\captionsetup[subfigure]{justification=justified,singlelinecheck=false}
    \centering
    \begin{subfigure}[t]{\linewidth}
        \centering
        \imagebox{37.5mm}{\includegraphics[width=\linewidth]{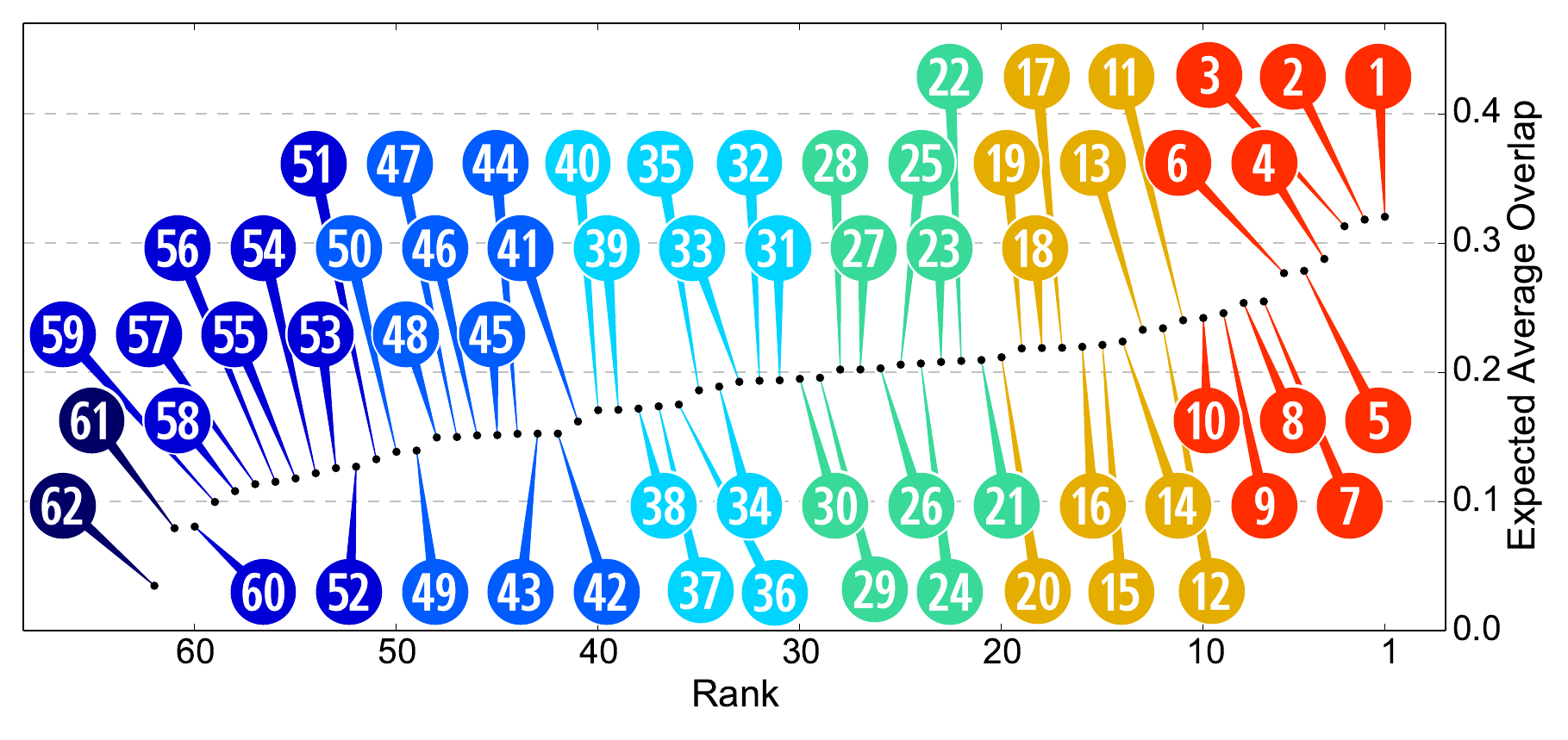}}
    \end{subfigure}%
    \vspace{\fill}
    \begin{subfigure}[t]{\linewidth}
        \centering
        \imagebox{35mm}{\includegraphics[width=\linewidth]{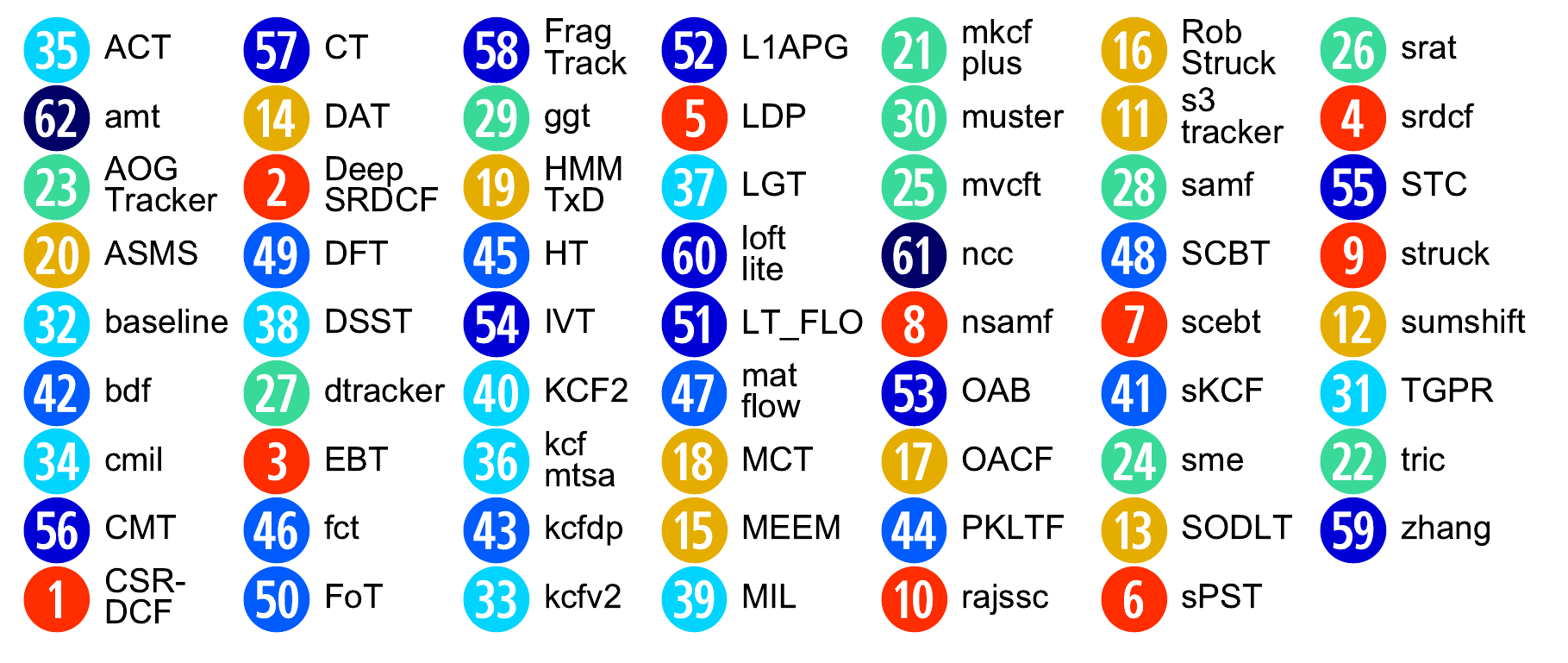}}
    \end{subfigure}
    \caption{Expected average overlap (EAO) plot for CSR-DSF (\#1) and all trackers participating in the VOT 2015~\citep{kristan_vot2015} benchmark listed below the plot in alphabetical order with their numerical codes.}
    \label{fig:vot15_graph}
\end{figure}

\begin{table}[!t]\setlength{\tabcolsep}{8pt}
\begin{center}
\caption{The ten top-performing trackers on the VOT2015 benchmark.}
\label{tab:detail-vot15}
\begin{tabular*}{0.9\linewidth}{l r r r}
\hline
Tracker & EAO & $A_\mathrm{av}$ & $R_\mathrm{av}$ \\ 
\hline
CSR-DCF & \first{0.320} & \third{0.55} & \second{0.93} \\ 
DeepSRDCF & \second{0.318} & \second{0.56} & \third{1.00} \\ 
EBT & \third{0.313} & 0.45 & \first{0.81} \\ 
srdcf & 0.288 & \third{0.55} & 1.18 \\ 
LDP & 0.278 & 0.49 & 1.30 \\ 
sPST & 0.277 & 0.54 & 1.42 \\ 
scebt & 0.255 & 0.54 & 1.72 \\ 
nsamf & 0.254 & 0.53 & 1.45 \\ 
struck & 0.246 & 0.46 & 1.50 \\ 
rajssc & 0.242 & \first{0.57} & 1.75 \\ 
\hline
\end{tabular*}
\end{center}
\end{table}

Figure~\ref{fig:vot15_graph} shows the VOT EAO plots with the CSR-DCF and the VOT2015 state-of-the-art approaches considering the VOT2016 rules that do not consider trackers learned on video sequences related to VOT to prevent over-fitting. The CSR-DCF outperforms all trackers and achieves the top rank. The CSR-DCF significantly outperforms the related correlation filter trackers like SRDCF~\citep{srdcf_iccv2015} as well as trackers that apply computationally-intesive state-of-the-art deep features e.g., deepSRDCF~\citep{danelljan_iccv2015_convolutional} and SO-DLT~\citep{wang_sodlt}. For completeness, detailed results for the ten top-performing trackers are shown in Table~\ref{tab:detail-vot15}.

\subsection{The VOT2016 benchmark~\citep{kristan_vot2016}} \label{sec:vot_2016}  

\begin{figure}[t!]
	\captionsetup[subfigure]{justification=justified,singlelinecheck=false}
    \centering
    \begin{subfigure}[t]{\linewidth}
        \centering
        \imagebox{37.5mm}{\includegraphics[width=\linewidth]{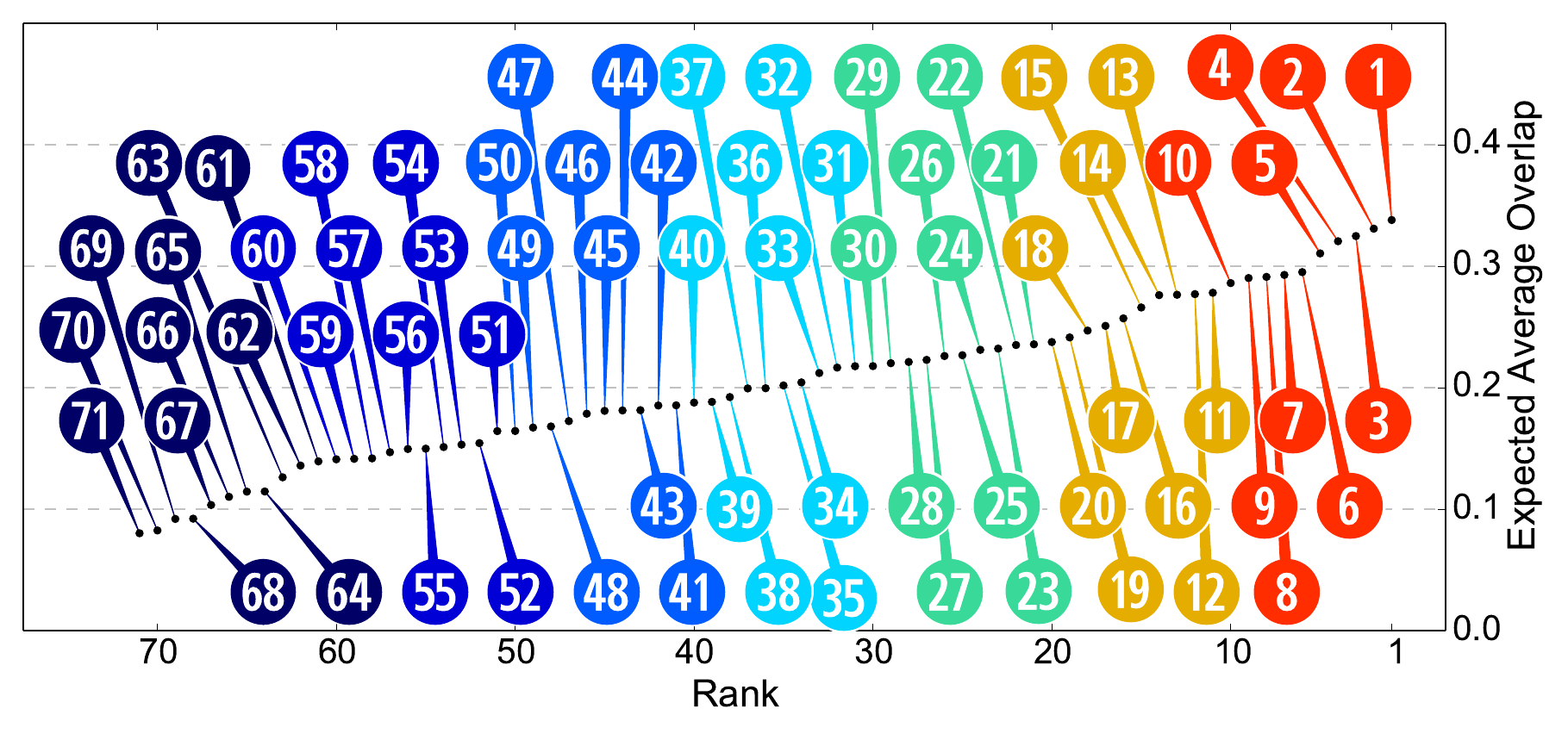}}
    \end{subfigure}%
    \vspace{\fill}
    \begin{subfigure}[t]{\linewidth}
        \centering
        \imagebox{45mm}{\includegraphics[width=\linewidth]{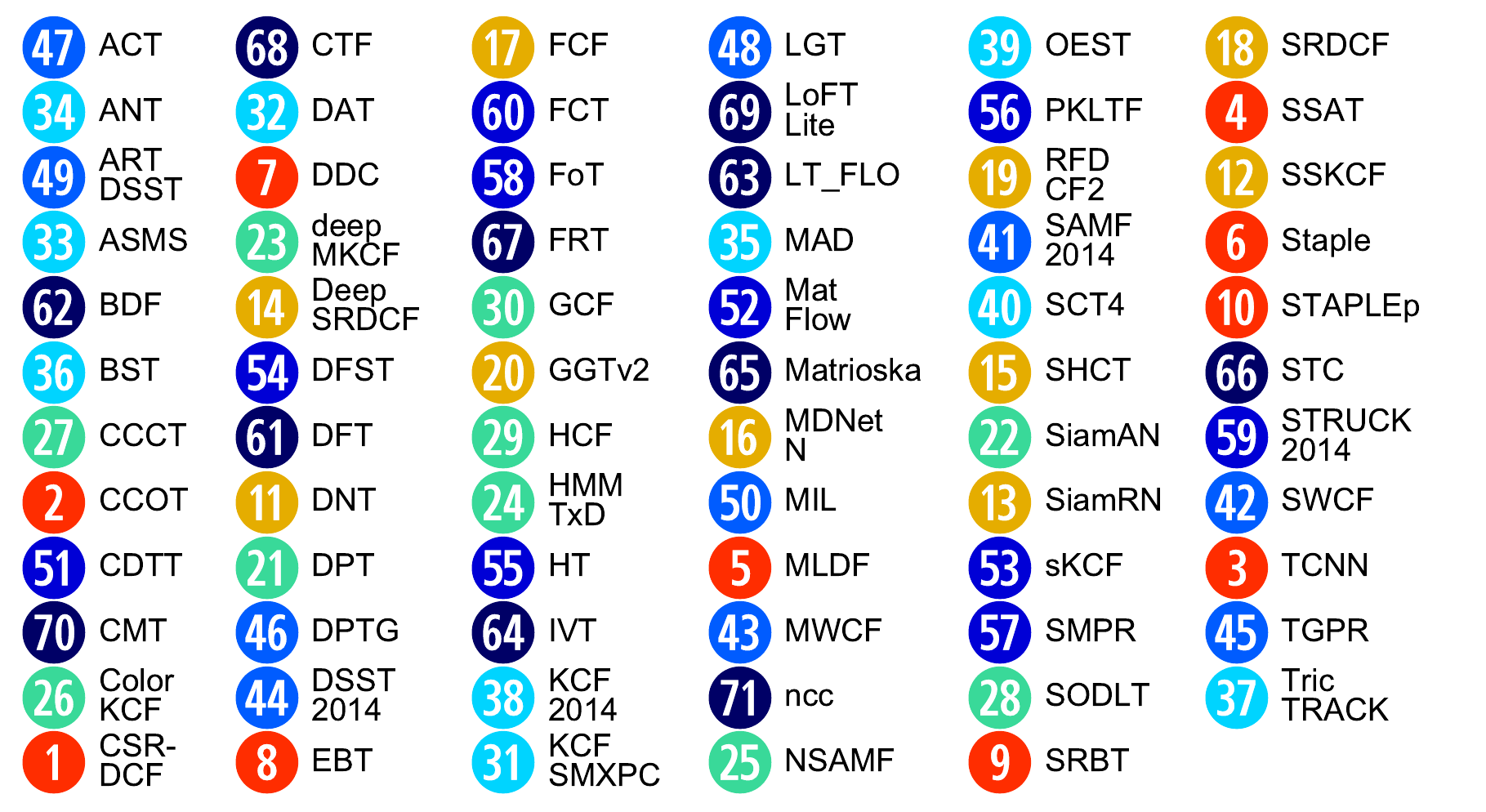}}
    \end{subfigure}
    \caption{Expected average overlap (EAO) plot for CSR-DSF (\#1) and all trackers participating in the VOT 2016~\citep{kristan_vot2016} benchmark listed below the plot in alphabetical order with their numerical codes.}
    \label{fig:vot16_graph}
\end{figure}

\begin{table}[!t]\setlength{\tabcolsep}{8pt}
\begin{center}
\caption{The ten top-performing trackers on the VOT2016 benchmark.}
\label{tab:detail-vot16}
\begin{tabular*}{0.9\linewidth}{l r r r}
\hline
Tracker & EAO & $A_\mathrm{av}$ & $R_\mathrm{av}$ \\ 
\hline
CSR-DCF & \first{0.338} & 0.51 & \second{0.85} \\ 
CCOT & \second{0.331} & 0.52 & \second{0.85} \\ 
TCNN & \third{0.325} & \third{0.54} & 0.96 \\ 
SSAT & 0.321 & \first{0.57} & 1.04 \\ 
MLDF & 0.311 & 0.48 & \first{0.83} \\ 
Staple & 0.295 & \third{0.54} & 1.35 \\ 
DDC & 0.293 & 0.53 & 1.23 \\ 
EBT & 0.291 & 0.44 & \third{0.90} \\ 
SRBT & 0.290 & 0.50 & 1.25 \\ 
STAPLEp & 0.286 & \second{0.55} & 1.32 \\ 
\hline
\end{tabular*}
\end{center}
\end{table}

Finally, we assess our tracker on the most recent visual tracking benchmark, VOT2016~\citep{kristan_vot2016}. The dataset contains 60 sequences from VOT2015~\citep{kristan_vot2015} with improved annotations. The benchmark evaluated a set of 70 trackers which includes the recently published and yet unpublished state-of-the-art trackers. The set is indeed diverse, the top-performing trackers come from various classes e.g., correlation filter methods: CCOT~\citep{danelljan_eccv2016_ccot}, Staple~\citep{staple_cvpr2016}, DDC~\citep{kristan_vot2016}, deep convolutional network based: TCNN~\citep{kristan_vot2016}, SSAT~\citep{kristan_vot2016,mdnet_cvpr2016}, MLDF~\citep{kristan_vot2016,wang_iccv2015}, FastSiamnet~\citep{siamese_arxiv} and different detection-based approaches: EBT~\citep{ebt_cvpr2016} and SRBT~\citep{kristan_vot2016}.

Figure~\ref{fig:vot16_graph} shows the EAO performance on the VOT2016. The CSR-DCF outperforms all 70 trackers with the EAO score equal to 0.338. The CSR-DCF significantly outperforms correlation filter approaches that do not apply deep ConvNets. Even though the CSR-DCF applies only simple features, it outperforms all trackers that apply computationally intensive deep features. Detailed performance scores for the ten top-performing trackers are shown in Table~\ref{tab:detail-vot16}.

\subsection{Per-attribute analysis} \label{sec:vot_2016_per_attribute} 

\begin{figure}[ht!]
\centering
\includegraphics[width=\linewidth]{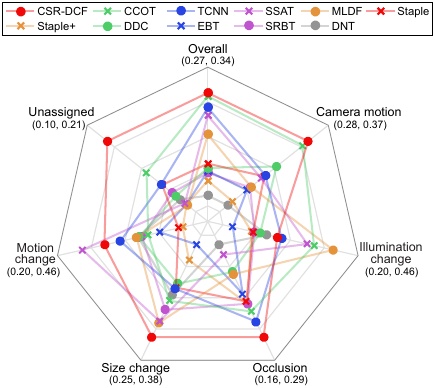}
\caption{Expected averaged overlap performance on different visual attributes on the VOT2016~\citep{kristan_vot2016} benchmark. The  CSR-DCF and the top 10 performing trackers from VOT2016 are shown. The scales of visual attribute axes are shown below the attribute labels.}
\label{fig:vot16_attributes}
\end{figure}

The VOT2016~\citep{kristan_vot2016} dataset is per-frame annotated with visual attributes and allows detailed analysis of per-attribute tracking performance. Figure~\ref{fig:vot16_attributes} shows per-attribute plot for ten top-performing trackers on VOT2016 in EAO. The  CSR-DCF is consistently ranked among top three trackers on five out of six attributes. In four attributes (size change, occlusion, camera motion, unassigned) the tracker is ranked top. The only attribute on which our CSR-DCF is outperformed by four of the compared trackers (MLDF, CCOT, SSAT and TCNN) is illumination change. All of these trackers use deep CNN features which are much more complex than the features in CSR-DCF and better handle illumination change.

\subsection{Tracking speed analysis} \label{sec:realtime_2016}
 
Tracking speed is an important factor of many real-world tracking problems. Table~\ref{tab:real-time} thus compares several related and well-known trackers (including the best-performing tracker on the VOT2016 challenge) in terms of speed and VOT performance measures. Speed measurements on a single CPU are computed on Intel Core i7 3.4GHz standard desktop. 

The  CSR-DCF performs on par with the VOT2016 best-performing CCOT~\citep{danelljan_eccv2016_ccot}, which applies deep ConvNets, with respect to VOT measures, while being 20 times faster than the CCOT. 
The CCOT was modified by replacing the computationally intensive deep features with the same simple features used in CSR-DCF. The resulting tracker, indicated by CCOT*, is still ten times slower than CSR-DCF, while the performance drops by over 15\%. The  CSR-DCF performs twice as fast as the related SRDCF~\citep{srdcf_iccv2015}, while achieving approximately $25\%$ better tracking results. 
The speed of baseline real-time trackers like DSST~\citep{danelljan2014accurate} and Struck~\citep{hare_struck} is comparable to CSR-DCF, but their tracking performance is significantly poorer. The fastest compared tracker, KCF~\citep{henriques2015tracking} runs much faster than real-time, but delivers a significantly poorer performance than CSR-DCF. 
 
The experiments show that the CSR-DCF tracks comparably to the state-of-the-art trackers which apply computationally demanding high-dimensional features, but runs considerably faster and delivers top tracking performance among the real-time trackers. 

\begin{figure}[!t]
\centering
\includegraphics[width=0.8\linewidth]{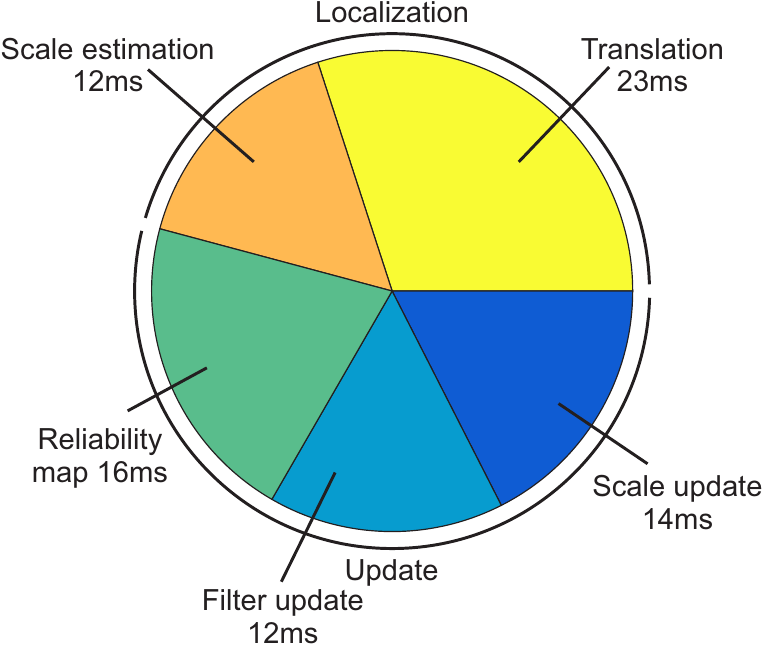}
\caption{A single iteration processing time decomposed across the main steps of the CSR-DCF.}
\label{fig:time_pie_chart}
\end{figure}

The average speed of our tracker measured on the VOT 2016 dataset is approximately 13 frames-per-second\footnote{With some basic code optimization and refactoring we speed-up our algorithm to 19 FPS without significant performance drop (only one additional failure on VOT2016 dataset).} or 77 milliseconds per-frame. Figure~\ref{fig:time_pie_chart} shows the processing time required by each step of the CSR-DCF. A tracking iteration is divided into two steps: (i) target localization and (ii) the visual model update. Target localization takes in average 35 milliseconds at each frame and is composed of two sub-steps: estimation of object translation (23ms) and scale change estimation (12ms). The visual model update step takes on average 42 milliseconds. It consists of three sub-steps: spatial reliability map estimation (16ms), filter update (12ms) and scale model update (14ms). Filter optimization, which is part of the filter update step, takes on average 7 milliseconds.

\begin{table}[!t]\setlength{\tabcolsep}{4pt}
\begin{center}
\caption{Speed in frames per second (fps) of correlation trackers and Struck -- a baseline. The EAO, average accuracy ($A_\mathrm{av}$) and average failures ($R_\mathrm{av}$) are shown for reference.}
\label{tab:real-time}
\begin{tabular*}{1\linewidth}{l l r r r r}
\hline
 \multicolumn{2}{c}{Tracker} & \multicolumn{1}{c}{EAO} & \multicolumn{1}{c}{$A_\mathrm{av}$} & \multicolumn{1}{c}{$R_\mathrm{av}$} & \multicolumn{1}{c}{fps} \\
\hline
\multicolumn{2}{l}{CSR-DCF} & \first{0.338} & \second{0.51} & \first{0.85} & \third{~~13.0} \\
CCOT & \tiny{ECCV2016} & \second{0.331} & \first{0.52} & \first{0.85} & 0.6 \\ 
CCOT* & \tiny{ECCV2016} & \third{0.274} & \first{0.52} & \second{1.18} & 1.0 \\
SRDCF & \tiny{ICCV2015} & 0.247 & \first{0.52} & \third{1.50} & 7.3 \\
KCF & \tiny{PAMI2015} & 0.192 & \third{0.48} & 2.03 & \first{115.7} \\
DSST & \tiny{PAMI2016} & 0.181 & \third{0.48} & 2.52 & \second{~~18.6} \\
Struck & \tiny{ICCV2011} & 0.142 & 0.42 & 3.37 & 8.5 \\
\hline
\end{tabular*}
\end{center}
\end{table}

\subsection{Qualitative evaluation} \label{sec:qualitative}

\begin{figure*}[!t]
\centering
\includegraphics[width=\linewidth]{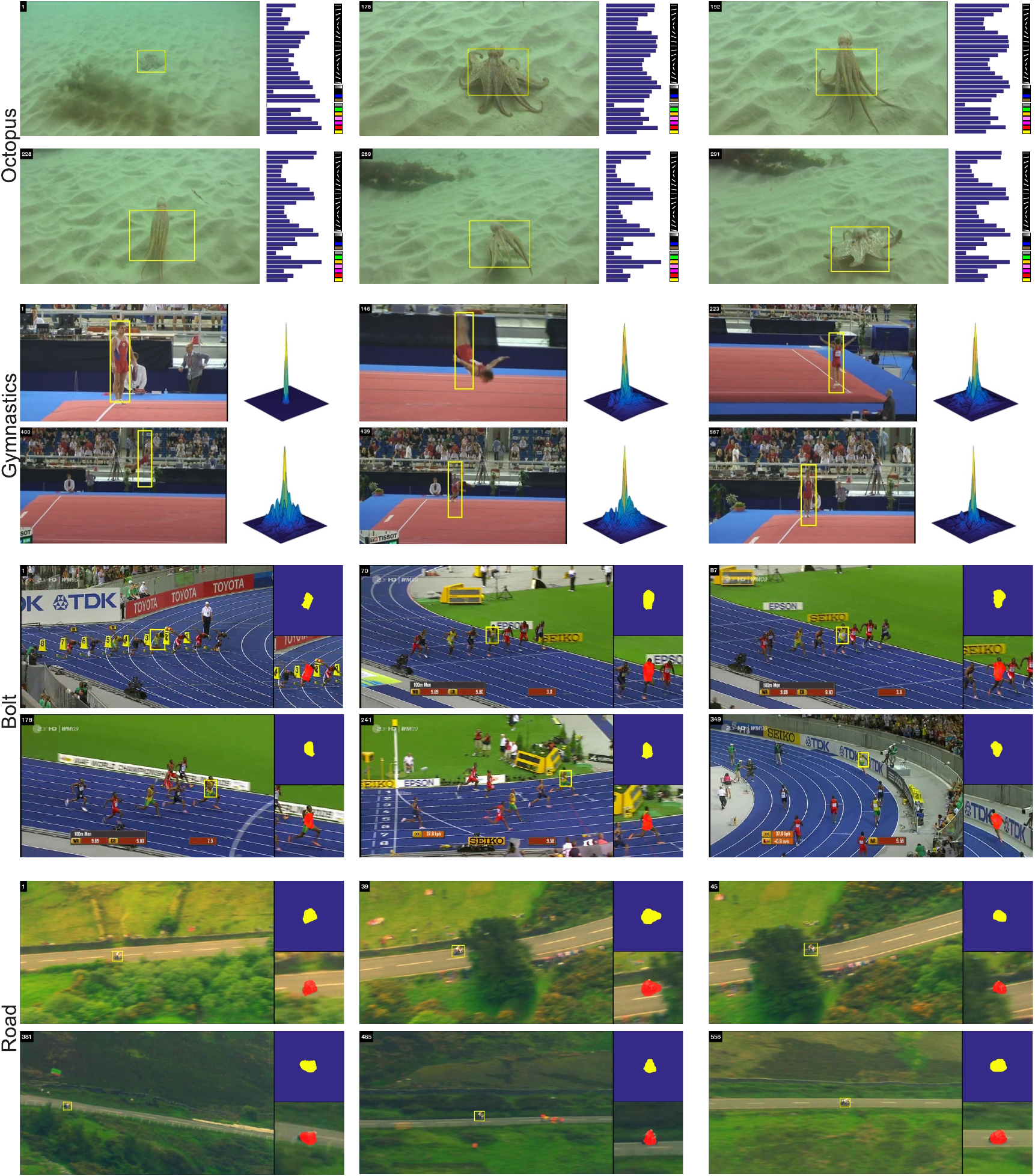}
\caption{Qualitative results of tracking with the CSR-DCF on four video sequences.}
\label{fig:qualitative}
\end{figure*}

Figure~\ref{fig:qualitative} shows four examples of tracking with the  CSR-DCF. In the following we describe tracking performance on each sequence.

The first example shows tracking of an octopus along with channel reliability weights. The first eighteen weights correspond to HoG channels, the $19\mathrm{th}$  weight is reliability of a grayscale template and the last ten weights correspond to colornames. Note that the colors in boxes are not the actual colors of the colornames, because these features are subspace of original colornames, designed to improve correlation filter tracking (see~\citep{danelljan2014adaptive}). Observe that when the octopus changes shape significantly, some channels become more discriminative than the others -- this is particularly evident in the first eighteen channels that represent the HoG features.

Tracking a gymnast is shown in the second example. The target is deforming and rotating over the sequence significantly, while our tracker is able to successfully track it. Additionally, the correlation response from the localization step is shown for each frame. The peak in the response is well expressed, which means that the filter accurately represents the target and that the discriminative channels overrule the less discriminative ones by our channel reliability estimation approach.

The third example shows tracking a sprinter. The spatial reliability map is visualized next to each frame. In the bottom-right corner of each frame the tracking patch is overlaid with the spatial reliability map. The reliability maps fit the target well and prevent the filter from learning the background.

The last example shows tracking under occlusion, i.e., a motorcyclist driving on the road while being repeatedly occluded by the trees. This example demonstrates that our tracker is robust to short-term full occlusions and that it is able to recover and localize the target despite the full occlusion. This is possible due to the robust learning with channel and spatial reliability map and the large capture range that our learning scheme provides.

\section{Conclusion}  \label{sec:conclusion}

The Discriminative Correlation Filter with Channel and Spatial Reliability (CSR-DCF) was introduced. The spatial reliability map adapts the filter support to the part of the object suitable for tracking which overcomes both the problems of circular shift enabling an arbitrary search region size and the limitations related to the rectangular shape assumption. A novel efficient spatial reliability map estimation 
method was proposed and an efficient optimization procedure was used for learning a correlation filter with the support constrained by the estimated map. The second novelty of CSR-DCF is the channel reliability. The reliability is estimated from the properties of the constrained least-squares solution. The channel reliability scores were used for weighting the per-channel filter responses in localization.

Experimental comparison with recent related state-of-the-art  boundary-constraints formulations showed significant benefits of using our formulation. The CSR-DCF achieves state-of-the-art performance on standard benchmarks: OTB100 \citep{otb_pami2015}, VOT2015 \citep{kristan_vot2015} and VOT2016 \citep{kristan_vot2016} while running close to the real-time on a single CPU. Despite using simple features like HoG and Colornames, the CSR-DCF performs on par with trackers that apply computationally complex deep ConvNet, but is significantly faster. 
 
To the best of our knowledge, the proposed approach is the first of its kind to introduce constrained filter learning with arbitrary spatial reliability map and the use of channel reliabilities. The spatial and channel reliability formulation is general and can be used in most modern correlation filters, e.g. those using deep features. 


\section{Appendix 1: Derivation of the augmented Lagrangian minimizer}  \label{sec:lagrangian_derivation}

This section provides a complete derivation of the relations (\ref{eq:min_hc}, \ref{eq:min_h}) in the Section~\ref{sec:constrained_learning}. The augmented Lagrangian from Equation~(\ref{eq:augmented_lagrange}) is
\begin{eqnarray}\label{eq:augmented_lagrange_orig}
	\mathcal{L}(\hat{\mathbf{h}}_c, \mathbf{h}, \hat{\mathbf{l}}) = \| \mathrm{diag}(\hat{\mathbf{f}}) \overline{\hat{\mathbf{h}}}_c - \hat{\mathbf{g}} \|^2 + \frac{\lambda}{2} \| \mathbf{h}_m \|^2 +\\ \nonumber
[\hat{\mathbf{l}}^H(  \hat{\mathbf{h}}_c - \hat{\mathbf{h}}_m) + \overline{\hat{\mathbf{l}}^H(  \hat{\mathbf{h}}_c -\hat{\mathbf{h}}_m)}] + \mu \|  \hat{\mathbf{h}}_c - \hat{\mathbf{h}}_m \|^2,
\end{eqnarray}
with $\mathbf{h}_m = (\mathbf{m} \odot \mathbf{h})$. For the purposes of derivation we will rewrite (\ref{eq:augmented_lagrange_orig}) into a fully vectorized form
\begin{gather}\label{eq:augmented_lagrange2}
	\mathcal{L}(\hat{\mathbf{h}}_c, \mathbf{h}, \hat{\mathbf{l}}) = \| \mathrm{diag}(\hat{\mathbf{f}}) \overline{\hat{\mathbf{h}}}_c - \hat{\mathbf{g}} \|^2 + \frac{\lambda}{2} \| \mathbf{h}_m \|^2 +\\ \nonumber
\bigg[\hat{\mathbf{l}}^H(  \hat{\mathbf{h}}_c - \sqrt{D}\mathbf{F}\mathbf{M}\mathbf{h}) + \overline{\hat{\mathbf{l}}^H(  \hat{\mathbf{h}}_c - \sqrt{D}\mathbf{F}\mathbf{M}\mathbf{h})}\bigg] + \\ \nonumber \mu \|  \hat{\mathbf{h}}_c - \sqrt{D}\mathbf{F}\mathbf{M}\mathbf{h} \|^2,
\end{gather}
where $\mathbf{F}$ denotes $D\times D$ orthonormal matrix of Fourier coefficients, such that the Fourier transform is defined as $\hat{\mathbf{x}} = \mathcal{F}(\mathbf{x}) = \sqrt{D}\mathbf{F}\mathbf{x}$ and $\mathbf{M}=\mathrm{diag}(\mathbf{m})$. For clearer representation we denote the four terms in the summation (\ref{eq:augmented_lagrange2}) as
\begin{equation}\label{eq:lagrange_terms}
	\mathcal{L}(\hat{\mathbf{h}}_c, \mathbf{h}, \hat{\mathbf{l}}) = \mathcal{L}_{1} + \mathcal{L}_{2} + \mathcal{L}_{3} + \mathcal{L}_{4},
\end{equation}
where 
\begin{eqnarray} \label{eq:hc_minimization00}
	 \mathcal{L}_1 = \overline{\Big( \mathrm{diag}(\hat{\mathbf{f}}) \overline{\hat{\mathbf{h}}}_c - \hat{\mathbf{g}} \Big)}^{T} \Big( \mathrm{diag}(\hat{\mathbf{f}}) \overline{\hat{\mathbf{h}}}_c - \hat{\mathbf{g}} \Big), 
\end{eqnarray}
\begin{eqnarray} \label{eq:hc_minimization01}
	 \mathcal{L}_2 = \frac{\lambda}{2} \| \mathbf{h}_m \|^2, 
\end{eqnarray}
\begin{eqnarray} \label{eq:hc_minimization02}
	 \mathcal{L}_3 = \hat{\mathbf{l}}^H(  \hat{\mathbf{h}}_c - \sqrt{D}\mathbf{F}\mathbf{M}\mathbf{h}) + \overline{\hat{\mathbf{l}}^H(  \hat{\mathbf{h}}_c - \sqrt{D}\mathbf{F}\mathbf{M}\mathbf{h})}, 
\end{eqnarray}
\begin{eqnarray} \label{eq:hc_minimization03}
	 \mathcal{L}_4 = \mu \|  \hat{\mathbf{h}}_c - \sqrt{D}\mathbf{F}\mathbf{M}\mathbf{h} \|^2.
\end{eqnarray}
Minimization of Equation~(\ref{eq:augmented_lagrange}) in Section~\ref{sec:constrained_learning} is an iterative process at which the following minimizations are required:
\begin{eqnarray}\label{eq:admm_min_}  
	\hat{\mathbf{h}}_c^\mathrm{opt} = \mathop {\arg \min }\limits_\mathbf{h_c} \mathcal{L}	(\hat{\mathbf{h}}_c, \mathbf{h}, \hat{\mathbf{l}}),\\
     \mathbf{h}^\mathrm{opt} = \mathop {\arg \min }\limits_\mathbf{h} \mathcal{L}			(\hat{\mathbf{h}}_c^\mathrm{opt}, \mathbf{h}, \hat{\mathbf{l}}). \label{eq:admm_min2}
\end{eqnarray}
Minimization w.r.t. to $\hat{\mathbf{h}}_c$ is derived by finding $\hat{\mathbf{h}}_c$ at which the complex gradient of the augmented Lagrangian vanishes, i.e.,
\begin{gather} \label{eq:hc_minimization1}
	\nabla_{\overline{\hat{\mathbf{h}}}_c} \mathcal{L} \equiv 0, \\	
	\nabla_{\overline{\hat{\mathbf{h}}}_c} \mathcal{L}_{1} + \nabla_{\overline{\hat{\mathbf{h}}}_c} \mathcal{L}_{2} + \nabla_{\overline{\hat{\mathbf{h}}}_c} \mathcal{L}_{3} + \nabla_{\overline{\hat{\mathbf{h}}}_c} \mathcal{L}_{4} \equiv 0. \label{eq:hc_minimization1_2}
\end{gather}
The partial complex gradients are:
\begin{gather} \label{eq:hc_minimization2}
	\nabla_{\overline{\hat{\mathbf{h}}}_c} \mathcal{L}_{1} = \\ \nonumber 
	= \frac{\partial}{\partial\overline{\hat{\mathbf{h}}}_c} \bigg[ \overline{\Big( \mathrm{diag}(\hat{\mathbf{f}}) \overline{\hat{\mathbf{h}}}_c - \hat{\mathbf{g}} \Big)}^{T} \Big( \mathrm{diag}(\hat{\mathbf{f}}) \overline{\hat{\mathbf{h}}}_c - \hat{\mathbf{g}} \Big) \bigg] = \\ \nonumber
	= \frac{\partial}{\partial\overline{\hat{\mathbf{h}}}_c} \bigg[
	\hat{\mathbf{h}}_c^T \mathrm{diag}(\hat{\mathbf{f}})^H \mathrm{diag}(\hat{\mathbf{f}}) \overline{\hat{\mathbf{h}}}_c - \hat{\mathbf{h}}_{c}^{T} \mathrm{diag}(\hat{\mathbf{f}})^{H} \hat{\mathbf{g}} - \\ \nonumber \hat{\mathbf{g}}^{H} \mathrm{diag}(\hat{\mathbf{f}}) \overline{\hat{\mathbf{h}}}_c + \hat{\mathbf{g}}^{H}\hat{\mathbf{g}} \bigg] = \\ \nonumber
	= \mathrm{diag}(\hat{\mathbf{f}})^{H} \mathrm{diag}(\hat{\mathbf{f}}) \hat{\mathbf{h}}_c - \mathrm{diag}(\hat{\mathbf{f}}) \overline{\hat{\mathbf{g}}},
\end{gather}

\begin{equation} \label{eq:hc_minimization3}
	\nabla_{\overline{\hat{\mathbf{h}}}_c} \mathcal{L}_{2} = 0,
\end{equation}

\begin{gather} \label{eq:hc_minimization4}
	\nabla_{\overline{\hat{\mathbf{h}}}_c} \mathcal{L}_{3} = \\ \nonumber 
	= \frac{\partial}{\partial\overline{\hat{\mathbf{h}}}_c} \bigg[ \hat{\mathbf{l}}^H \Big( \hat{\mathbf{h}}_c - \sqrt{D}\mathbf{F}\mathbf{M}\mathbf{h} \Big) +  \overline{\hat{\mathbf{l}}^H \Big( \hat{\mathbf{h}}_c - \sqrt{D}\mathbf{F}\mathbf{M}\mathbf{h} \Big)} \bigg] = \\ \nonumber
	= \frac{\partial}{\partial\overline{\hat{\mathbf{h}}}_c} \bigg[ \hat{\mathbf{l}}^H \hat{\mathbf{h}}_c - \hat{\mathbf{l}}^H \sqrt{D}\mathbf{F}\mathbf{M}\mathbf{h} + \hat{\mathbf{l}}^T \overline{\hat{\mathbf{h}}}_c - \hat{\mathbf{l}}^T \sqrt{D}\overline{\mathbf{F}}\overline{\mathbf{M}}\overline{\mathbf{h}} \bigg] = \\ \nonumber = \hat{\mathbf{l}},
\end{gather}

\begin{gather} \label{eq:hc_minimization5}
	\nabla_{\overline{\hat{\mathbf{h}}}_c} \mathcal{L}_{4} = \\ \nonumber
	= \frac{\partial}{\partial\overline{\hat{\mathbf{h}}}_c} \bigg[ \mu \overline{\Big( \hat{\mathbf{h}}_c - \sqrt{D}\mathbf{F}\mathbf{M}\mathbf{h} \Big)}^{T} \Big( \hat{\mathbf{h}}_c - \sqrt{D}\mathbf{F}\mathbf{M}\mathbf{h} \Big) \bigg] = \\ \nonumber
	= \frac{\partial}{\partial\overline{\hat{\mathbf{h}}}_c} \bigg[ \mu \Big( \hat{\mathbf{h}}_c^{H} \hat{\mathbf{h}}_c - \hat{\mathbf{h}}_c^{H}\sqrt{D}\mathbf{F}\mathbf{M}\mathbf{h} - \\ \nonumber \sqrt{D}\mathbf{h}^T\mathbf{M}\mathbf{F}^{H}\hat{\mathbf{h}}_c + D\mathbf{h}^T\mathbf{M}\mathbf{F}^{H}\mathbf{F}\mathbf{M}\mathbf{h} \Big) \bigg] = \\ \nonumber
	= \mu \hat{\mathbf{h}}_c - \mu \sqrt{D}\mathbf{F}\mathbf{M}\mathbf{h}.
\end{gather}
Note that $\sqrt{D}\mathbf{F}\mathbf{M}\mathbf{h} = \hat{\mathbf{h}}_m$ according to our original definition of $\hat{\mathbf{h}}_m$. Plugging (\ref{eq:hc_minimization2}-\ref{eq:hc_minimization5}) into (\ref{eq:hc_minimization1_2}) yields
\begin{gather} \label{eq:hc_minimization6}
	\mathrm{diag}(\hat{\mathbf{f}})^{H} \mathrm{diag}(\hat{\mathbf{f}}) \hat{\mathbf{h}}_c - \mathrm{diag}(\hat{\mathbf{f}}) \overline{\hat{\mathbf{g}}} + \hat{\mathbf{l}} + \mu \hat{\mathbf{h}}_c - \mu \hat{\mathbf{h}}_m = 0, \\ \nonumber
	\hat{\mathbf{h}}_c = \frac{\mathrm{diag}(\hat{\mathbf{f}}) \overline{\hat{\mathbf{g}}} + \mu \hat{\mathbf{h}}_m - \hat{\mathbf{l}}}{\mathrm{diag}(\hat{\mathbf{f}})^{H} \mathrm{diag}(\hat{\mathbf{f}}) + \mu},
\end{gather}
which can be rewritten into
\begin{gather} \label{eq:hc_minimization7}
	\hat{\mathbf{h}}_c = \frac{\hat{\mathbf{f}} \odot \overline{\hat{\mathbf{g}}} + \mu \hat{\mathbf{h}}_m - \hat{\mathbf{l}}}{\overline{\hat{\mathbf{f}}} \odot \hat{\mathbf{f}} + \mu}.
\end{gather}
Next we derive the closed-form solution of~(\ref{eq:admm_min2}). The optimal $\mathbf{h}$ is obtained when the complex gradient w.r.t. $\mathbf{h}$ vanishes, i.e.,

\begin{gather} \label{eq:h_minimization1}
	\nabla_{\overline{\mathbf{h}}} \mathcal{L} \equiv 0 
 \\	
	\nabla_{\overline{\mathbf{h}}} \mathcal{L}_{1} + \nabla_{\overline{\mathbf{h}}} \mathcal{L}_{2} + \nabla_{\overline{\mathbf{h}}} \mathcal{L}_{3} + \nabla_{\overline{\mathbf{h}}} \mathcal{L}_{4} \equiv 0. \label{eq:h_minimization1_2}
\end{gather}
The partial gradients are 
\begin{equation} \label{eq:h_minimization2}
	\nabla_{\overline{\mathbf{h}}} \mathcal{L}_{1} = 0,
\end{equation}
\begin{gather} \label{eq:h_minimization3}
	\nabla_{\overline{\mathbf{h}}} \mathcal{L}_{2} = \\ \nonumber
	= \frac{\partial}{\partial\overline{\mathbf{h}}} \bigg[ \frac{\lambda}{2}  \overline{(\mathbf{M} \mathbf{h})}^T (\mathbf{M} \mathbf{h}) \bigg] = \frac{\partial}{\partial\overline{\mathbf{h}}} \bigg[ \frac{\lambda}{2}  \mathbf{h}^H \overline{\mathbf{M}} \mathbf{M} \mathbf{h} \bigg].
\end{gather}
Since we defined mask $\mathbf{m}$ as a binary mask, the product $\overline{\mathbf{M}} \mathbf{M}$ can be simplified into $\mathbf{M}$ and the result for $\nabla_{\overline{\mathbf{h}}} \mathcal{L}_{2}$ is
\begin{gather} \label{eq:h_minimization4}
	\nabla_{\overline{\mathbf{h}}} \mathcal{L}_{2} = \frac{\lambda}{2} \mathbf{M} \mathbf{h}.
\end{gather}
The remaining gradients are as follows:
\begin{gather} \label{eq:h_minimization5}
	\nabla_{\overline{\mathbf{h}}} \mathcal{L}_{3} = \\ \nonumber
	= \frac{\partial}{\partial\overline{\mathbf{h}}} \bigg[ \hat{\mathbf{l}}^H \Big( \hat{\mathbf{h}}_c - \sqrt{D}\mathbf{F}\mathbf{M}\mathbf{h} \Big) + \overline{\hat{\mathbf{l}}^H \Big( \hat{\mathbf{h}}_c - \sqrt{D}\mathbf{F}\mathbf{M}\mathbf{h} \Big)} \bigg] = \\ \nonumber
	= \frac{\partial}{\partial\overline{\mathbf{h}}} \bigg[ \hat{\mathbf{l}}^H \hat{\mathbf{h}}_c - \hat{\mathbf{l}}^H \sqrt{D}\mathbf{F}\mathbf{M}\mathbf{h} + \hat{\mathbf{l}}^T \overline{\hat{\mathbf{h}}}_c - \hat{\mathbf{l}}^T \sqrt{D}\overline{\mathbf{F}}\overline{\mathbf{M}}\overline{\mathbf{h}} \bigg] = \\ \nonumber
	= - \sqrt{D} \mathbf{M} \mathbf{F}^H \hat{\mathbf{l}},
\end{gather}

\begin{gather} \label{eq:h_minimization6}
	\nabla_{\overline{\mathbf{h}}} \mathcal{L}_{4} = \\ \nonumber
	= \frac{\partial}{\partial\overline{\mathbf{h}}} \bigg[ \mu \overline{\Big( \hat{\mathbf{h}}_c - \sqrt{D}\mathbf{F}\mathbf{M}\mathbf{h} \Big)}^{T} \Big( \hat{\mathbf{h}}_c - \sqrt{D}\mathbf{F}\mathbf{M}\mathbf{h} \Big) \bigg] = \\ \nonumber
	= \frac{\partial}{\partial\overline{\mathbf{h}}} \bigg[ \mu \Big( \hat{\mathbf{h}}_c^H \hat{\mathbf{h}}_c - \hat{\mathbf{h}}_c^H\sqrt{D}\mathbf{F}\mathbf{M}\mathbf{h} - \\ \nonumber \sqrt{D}\mathbf{h}^H\mathbf{M}\mathbf{F}^H\hat{\mathbf{h}}_c + D\mathbf{h}^H \mathbf{M}\mathbf{h} \Big) \bigg] = \\ \nonumber
	= - \mu \sqrt{D}\mathbf{M}\mathbf{F}^H \hat{\mathbf{h}}_c + \mu D\mathbf{M}\mathbf{h}.
\end{gather}
Plugging~(\ref{eq:h_minimization2}-\ref{eq:h_minimization6}) into~(\ref{eq:h_minimization1_2}) yields 
\begin{gather} \label{eq:h_minimization7}
	\frac{\lambda}{2} \mathbf{M} \mathbf{h} - \sqrt{D} \mathbf{M} \mathbf{F}^H \hat{\mathbf{l}} - \mu \sqrt{D}\mathbf{M}\mathbf{F}^H \hat{\mathbf{h}}_c + \mu D\mathbf{M}\mathbf{h} = 0, \\ \nonumber
	\mathbf{M}\mathbf{h} = \mathbf{M}\frac{\sqrt{D}\mathbf{F}^H(\hat{\mathbf{l}} + \mu \hat{\mathbf{h}}_c)}{\frac{\lambda}{2} + \mu D}.
\end{gather}
Using the definition of the inverse Fourier transform, i.e., $\mathcal{F}^{-1}(\hat{\mathbf{x}}) = \frac{1}{\sqrt{D}} \mathbf{F}^H \hat{\mathbf{x}}$, (\ref{eq:h_minimization7}) can be rewritten into
\begin{gather} \label{eq:h_minimization8}
	\mathbf{m} \odot\mathbf{h} = \mathbf{m} \odot\frac{\mathcal{F}^{-1}(\hat{\mathbf{l}} + \mu \hat{\mathbf{h}}_c)}{\frac{\lambda}{2D} + \mu}.
\end{gather}
The values in $\mathbf{m}$ are either zero or one. Elements in $\mathbf{h}$ that correspond to the zeros in $\mathbf{m}$ can in principle not be recovered from (\ref{eq:h_minimization8}) since this would result in division by zero. But our initial definition of the problem was to seek solutions for the filter that satisfies the following relation $\mathbf{h} \equiv \mathbf{h} \odot \mathbf{m}$. This means the values corresponding to zeros in $\mathbf{m}$ should be zero in $\mathbf{h}$. Thus the proximal solution to~(\ref{eq:h_minimization8}) is 
\begin{gather} \label{eq:h_minimization9}
	\mathbf{h} = \mathbf{m} \odot \frac{\mathcal{F}^{-1}(\hat{\mathbf{l}} + \mu \hat{\mathbf{h}}_c)}{\frac{\lambda}{2D} + \mu}.
\end{gather}

\section*{Acknownledgements}
This work was supported in part by the following research programs and projects: Slovenian research agency research programs and projects P2-0214, L2-6765 and J2-8175. Ji\v{r}i Matas and Tom{\'a}\v{s} Voj\'{i}\~{r} were supported by The Czech Science Foundation Project GACR P103/12/G084 and Toyota Motor Europe. We would also like to thank dr. Rok Žitko for discussion on complex differentiation.


\bibliographystyle{spbasic}
\bibliography{bib}

\begin{thebibliography}{54}
\providecommand{\natexlab}[1]{#1}
\providecommand{\url}[1]{{#1}}
\providecommand{\urlprefix}{URL }
\expandafter\ifx\csname urlstyle\endcsname\relax
  \providecommand{\doi}[1]{DOI~\discretionary{}{}{}#1}\else
  \providecommand{\doi}{DOI~\discretionary{}{}{}\begingroup
  \urlstyle{rm}\Url}\fi
\providecommand{\eprint}[2][]{\url{#2}}

\bibitem[{Babenko et~al(2011)Babenko, Yang, and Belongie}]{babenko_mil}
Babenko B, Yang MH, Belongie S (2011) Robust object tracking with online
  multiple instance learning. IEEE Trans Pattern Anal Mach Intell
  33(8):1619--1632

\bibitem[{Bertinetto et~al(2016{\natexlab{a}})Bertinetto, Valmadre, Golodetz,
  Miksik, and Torr}]{staple_cvpr2016}
Bertinetto L, Valmadre J, Golodetz S, Miksik O, Torr PHS (2016{\natexlab{a}})
  Staple: Complementary learners for real-time tracking. In: Comp. Vis. Patt.
  Recognition, pp 1401--1409

\bibitem[{Bertinetto et~al(2016{\natexlab{b}})Bertinetto, Valmadre, Henriques,
  Vedaldi, and Torr}]{siamese_arxiv}
Bertinetto L, Valmadre J, Henriques JF, Vedaldi A, Torr PH (2016{\natexlab{b}})
  Fully-convolutional siamese networks for object tracking. arXiv preprint
  arXiv:160609549

\bibitem[{Bolme et~al(2010)Bolme, Beveridge, Draper, and Lui}]{bolme2010visual}
Bolme DS, Beveridge JR, Draper BA, Lui YM (2010) Visual object tracking using
  adaptive correlation filters. In: Comp. Vis. Patt. Recognition, IEEE, pp
  2544--2550

\bibitem[{Boyd et~al(2011)Boyd, Parikh, Chu, Peleato, and
  Eckstein}]{admm_boyd2011}
Boyd S, Parikh N, Chu E, Peleato B, Eckstein J (2011) Distributed optimization
  and statistical learning via the alternating direction method of multipliers.
  Foundations and Trends in Machine Learning 3(1):1--122

\bibitem[{{\v{C}}ehovin et~al(2016){\v{C}}ehovin, Leonardis, and
  Kristan}]{cehovin_tip2016}
{\v{C}}ehovin L, Leonardis A, Kristan M (2016) Visual object tracking
  performance measures revisited. IEEE Trans Image Proc 25(3):1261--1274

\bibitem[{Dalal and Triggs(2005)}]{dalal_triggs_hog}
Dalal N, Triggs B (2005) Histograms of oriented gradients for human detection.
  In: Comp. Vis. Patt. Recognition, vol~1, pp 886--893

\bibitem[{Danelljan et~al(2014{\natexlab{a}})Danelljan, H{\"a}ger, Khan, and
  Felsberg}]{danelljan2014accurate}
Danelljan M, H{\"a}ger G, Khan FS, Felsberg M (2014{\natexlab{a}}) Accurate
  scale estimation for robust visual tracking. In: Proc. British Machine Vision
  Conference, pp 1--11

\bibitem[{Danelljan et~al(2014{\natexlab{b}})Danelljan, Khan, Felsberg, and
  van~de Weijer}]{danelljan2014adaptive}
Danelljan M, Khan FS, Felsberg M, van~de Weijer J (2014{\natexlab{b}}) Adaptive
  color attributes for real-time visual tracking. In: 2014 {IEEE} Conference on
  Computer Vision and Pattern Recognition, {CVPR} 2014, Columbus, OH, USA, June
  23-28, 2014, pp 1090--1097

\bibitem[{Danelljan et~al(2015{\natexlab{a}})Danelljan, Hager, Shahbaz~Khan,
  and Felsberg}]{srdcf_iccv2015}
Danelljan M, Hager G, Shahbaz~Khan F, Felsberg M (2015{\natexlab{a}}) Learning
  spatially regularized correlation filters for visual tracking. In: Int. Conf.
  Computer Vision, pp 4310--4318

\bibitem[{Danelljan et~al(2015{\natexlab{b}})Danelljan, Häger, Khan, and
  Felsberg}]{danelljan_iccv2015_convolutional}
Danelljan M, Häger G, Khan FS, Felsberg M (2015{\natexlab{b}}) Convolutional
  features for correlation filter based visual tracking. In: IEEE International
  Conference on Computer Vision Workshop (ICCVW), pp 621--629

\bibitem[{Danelljan et~al(2016)Danelljan, Robinson, Khan, and
  Felsberg}]{danelljan_eccv2016_ccot}
Danelljan M, Robinson A, Khan FS, Felsberg M (2016) Beyond correlation filters:
  learning continuous convolution operators for visual tracking. In: Proc.
  European Conf. Computer Vision, Springer, pp 472--488

\bibitem[{Danelljan et~al(2017)Danelljan, Häger, Khan, and
  Felsberg}]{danelljan_dsst_pami}
Danelljan M, Häger G, Khan FS, Felsberg M (2017) Discriminative scale space
  tracking. IEEE Trans Pattern Anal Mach Intell 39(8):1561--1575

\bibitem[{Dinh et~al(2011)Dinh, Vo, and Medioni}]{cxt_cvpr2011}
Dinh TB, Vo N, Medioni G (2011) Context tracker: Exploring supporters and
  distracters in unconstrained environments. In: Comp. Vis. Patt. Recognition,
  pp 1177--1184

\bibitem[{Diplaros et~al(2007)Diplaros, Vlassis, and
  Gevers}]{diplaros_genmodel}
Diplaros A, Vlassis N, Gevers T (2007) A spatially constrained generative model
  and an em algorithm for image segmentation. IEEE Trans Neural Networks
  18(3):798 -- 808

\bibitem[{Felzenszwalb et~al(2010)Felzenszwalb, Girshick, McAllester, and
  Ramanan}]{felzenszwalb_dpm}
Felzenszwalb P, Girshick R, McAllester D, Ramanan D (2010) Object detection
  with discriminatively trained part-based models. IEEE Trans Pattern Anal Mach
  Intell 32(9):1627--1645

\bibitem[{Galoogahi et~al(2013)Galoogahi, Sim, and
  Lucey}]{galoogahi_multi_channel_correlation}
Galoogahi HK, Sim T, Lucey S (2013) Multi-channel correlation filters. In: Int.
  Conf. Computer Vision, pp 3072--3079

\bibitem[{Grabner et~al(2006)Grabner, Grabner, and Bischof}]{grabner_oab}
Grabner H, Grabner M, Bischof H (2006) Real-time tracking via on-line boosting.
  In: Proc. British Machine Vision Conference, vol~1, pp 47--56

\bibitem[{Hare et~al(2011)Hare, Saffari, and Torr}]{hare_struck}
Hare S, Saffari A, Torr PHS (2011) Struck: Structured output tracking with
  kernels. In: Int. Conf. Computer Vision, IEEE Computer Society, Washington,
  DC, USA, pp 263--270

\bibitem[{Henriques et~al(2012)Henriques, Caseiro, Martins, and
  Batista}]{csk_henriques_eccv2012}
Henriques JF, Caseiro R, Martins P, Batista J (2012) Exploiting the circulant
  structure of tracking-by-detection with kernels. In: Fitzgibbon A, Lazebnik
  S, Perona P, Sato Y, Schmid C (eds) Proc. European Conf. Computer Vision,
  Springer Berlin Heidelberg, Berlin, Heidelberg, pp 702--715

\bibitem[{Henriques et~al(2015)Henriques, Caseiro, Martins, and
  Batista}]{henriques2015tracking}
Henriques JF, Caseiro R, Martins P, Batista J (2015) High-speed tracking with
  kernelized correlation filters. IEEE Trans Pattern Anal Mach Intell
  37(3):583--596

\bibitem[{Hester and Casasent(1980)}]{hester1980}
Hester CF, Casasent D (1980) Multivariant technique for multiclass pattern
  recognition. Applied Optics 19(11):1758--1761

\bibitem[{Hong et~al(2015)Hong, Chen, Wang, Mei, Prokhorov, and
  Tao}]{muster_cvpr2015}
Hong Z, Chen Z, Wang C, Mei X, Prokhorov D, Tao D (2015) Multi-store tracker
  (muster): A cognitive psychology inspired approach to object tracking. In:
  Comp. Vis. Patt. Recognition, pp 749--758

\bibitem[{Kalal et~al(2012)Kalal, Mikolajczyk, and Matas}]{kalal_pami}
Kalal Z, Mikolajczyk K, Matas J (2012) Tracking-learning-detection. IEEE Trans
  Pattern Anal Mach Intell 34(7):1409--1422

\bibitem[{Kiani~Galoogahi et~al(2015)Kiani~Galoogahi, Sim, and
  Lucey}]{cfwlb_cvpr2015}
Kiani~Galoogahi H, Sim T, Lucey S (2015) Correlation filters with limited
  boundaries. In: Comp. Vis. Patt. Recognition, pp 4630--4638

\bibitem[{Kristan et~al(2013)Kristan, Pflugfelder, Leonardis, Matas, Porikli,
  Čehovin, Nebehay, Fernandez, and Vojir}]{kristan_vot2013}
Kristan M, Pflugfelder R, Leonardis A, Matas J, Porikli F, Čehovin L, Nebehay
  G, Fernandez G, Vojir Tea (2013) The visual object tracking vot2013 challenge
  results. In: Vis. Obj. Track. Challenge VOT2013, In conjunction with
  ICCV2013, pp 98--111

\bibitem[{Kristan et~al(2014)Kristan, Pflugfelder, Leonardis, Matas,
  \v{C}ehovin, Nebehay, Vojir, and et~al. Fernandez}]{kristan_vot2014}
Kristan M, Pflugfelder R, Leonardis A, Matas J, \v{C}ehovin L, Nebehay G, Vojir
  T, et~al Fernandez G (2014) The visual object tracking vot2014 challenge
  results. In: Proc. European Conf. Computer Vision, pp 191--217

\bibitem[{Kristan et~al(2015)Kristan, Matas, Leonardis, Felsberg, \v{C}ehovin,
  Fernandez, Vojir, H\"{a}ger, Nebehay, and et~al.
  Pflugfelder}]{kristan_vot2015}
Kristan M, Matas J, Leonardis A, Felsberg M, \v{C}ehovin L, Fernandez G, Vojir
  T, H\"{a}ger G, Nebehay G, et~al Pflugfelder R (2015) The visual object
  tracking vot2015 challenge results. In: Int. Conf. Computer Vision

\bibitem[{Kristan et~al(2016{\natexlab{a}})Kristan, Kenk, Kovačič, and
  Perš}]{kristan_tcyb2016}
Kristan M, Kenk VS, Kovačič S, Perš J (2016{\natexlab{a}}) Fast image-based
  obstacle detection from unmanned surface vehicles. IEEE Transactions on
  Cybernetics 46(3):641--654

\bibitem[{Kristan et~al(2016{\natexlab{b}})Kristan, Leonardis, Matas, Felsberg,
  Pflugfelder, \v{C}ehovin, Vojir, H\"{a}ger, Lukežič, and et~al.
  Fernandez}]{kristan_vot2016}
Kristan M, Leonardis A, Matas J, Felsberg M, Pflugfelder R, \v{C}ehovin L,
  Vojir T, H\"{a}ger G, Lukežič A, et~al Fernandez G (2016{\natexlab{b}}) The
  visual object tracking vot2016 challenge results. In: Proc. European Conf.
  Computer Vision

\bibitem[{Kristan et~al(2016{\natexlab{c}})Kristan, Matas, Leonardis, Vojir,
  Pflugfelder, Fernandez, Nebehay, Porikli, and
  Cehovin}]{kristan_vot_tpami2016}
Kristan M, Matas J, Leonardis A, Vojir T, Pflugfelder R, Fernandez G, Nebehay
  G, Porikli F, Cehovin L (2016{\natexlab{c}}) A novel performance evaluation
  methodology for single-target trackers. IEEE Trans Pattern Anal Mach Intell

\bibitem[{Li and Zhu(2014{\natexlab{a}})}]{samf_eccv2014}
Li Y, Zhu J (2014{\natexlab{a}}) A scale adaptive kernel correlation filter
  tracker with feature integration. In: Proc. European Conf. Computer Vision,
  pp 254--265

\bibitem[{Li and Zhu(2014{\natexlab{b}})}]{Li2014}
Li Y, Zhu J (2014{\natexlab{b}}) A scale adaptive kernel correlation filter
  tracker with feature integration. In: Proc. European Conf. Computer Vision,
  pp 254--265

\bibitem[{Liang et~al(2015)Liang, Blasch, and Ling}]{templecolor_tip2015}
Liang P, Blasch E, Ling H (2015) Encoding color information for visual
  tracking: Algorithms and benchmark. IEEE Trans Image Proc 24(12):5630--5644

\bibitem[{Liu et~al(2011)Liu, Huang, Yang, and Kulikowsk}]{lsk_cvpr2011}
Liu B, Huang J, Yang L, Kulikowsk C (2011) Robust tracking using local sparse
  appearance model and k-selection. In: Comp. Vis. Patt. Recognition, pp
  1313--1320

\bibitem[{Liu et~al(2016)Liu, Zhang, Cao, and Xu}]{structural_cf_cvpr2016}
Liu S, Zhang T, Cao X, Xu C (2016) Structural correlation filter for robust
  visual tracking. In: Comp. Vis. Patt. Recognition, pp 4312--4320

\bibitem[{Liu et~al(2015)Liu, Wang, and Yang}]{part_cf_cvpr2016}
Liu T, Wang G, Yang Q (2015) Real-time part-based visual tracking via adaptive
  correlation filters. In: Comp. Vis. Patt. Recognition, pp 4902--4912

\bibitem[{Luke\v{z}i\v{c} et~al(2017)Luke\v{z}i\v{c}, \v{C}. Zajc, and
  Kristan}]{lukezic_dpt}
Luke\v{z}i\v{c} A, \v{C} Zajc L, Kristan M (2017) Deformable parts correlation
  filters for robust visual tracking. IEEE Transactions on Cybernetics
  PP(99):1--13

\bibitem[{Ma et~al(2015)Ma, Huang, Yang, and
  Yang}]{convolutional_mingsungyang_iccv2015}
Ma C, Huang JB, Yang X, Yang MH (2015) Hierarchical convolutional features for
  visual tracking. In: Int. Conf. Computer Vision, pp 3074--3082

\bibitem[{Mueller et~al(2016)Mueller, Smith, and
  Ghanem}]{uav_benchmark_eccv2016}
Mueller M, Smith N, Ghanem B (2016) A benchmark and simulator for uav tracking.
  In: Proc. European Conf. Computer Vision

\bibitem[{Nam and Han(2016)}]{mdnet_cvpr2016}
Nam H, Han B (2016) Learning multi-domain convolutional neural networks for
  visual tracking. In: Comp. Vis. Patt. Recognition, pp 4293--4302

\bibitem[{Qi et~al(2016)Qi, Zhang, Qin, Yao, Huang, Lim, and
  Yang}]{qi_hedge_deep_cf}
Qi Y, Zhang S, Qin L, Yao H, Huang Q, Lim J, Yang MH (2016) Hedged deep
  tracking. In: CVPR, pp 4303--4311

\bibitem[{Smeulders et~al(2014)Smeulders, Chu, Cucchiara, Calderara, Dehghan,
  and Shah}]{alov_pami2014}
Smeulders A, Chu D, Cucchiara R, Calderara S, Dehghan A, Shah M (2014) Visual
  tracking: An experimental survey. IEEE Trans Pattern Anal Mach Intell
  36(7):1442--1468

\bibitem[{Vojir and Matas(2017)}]{vojir_segmentation}
Vojir T, Matas J (2017) Pixel-wise object segmentations for the {VOT} 2016
  dataset. Research Report CTU--CMP--2017--01, Center for Machine Perception,
  K13133 FEE Czech Technical University, Prague, Czech Republic

\bibitem[{Wang et~al(2015{\natexlab{a}})Wang, Ouyang, Wang, and
  Lu}]{wang_iccv2015}
Wang L, Ouyang W, Wang X, Lu H (2015{\natexlab{a}}) Visual tracking with fully
  convolutional networks. In: Int. Conf. Computer Vision, pp 3119--3127

\bibitem[{Wang et~al(2015{\natexlab{b}})Wang, Li, Gupta, and
  Yeung}]{wang_sodlt}
Wang N, Li S, Gupta A, Yeung D (2015{\natexlab{b}}) Transferring rich feature
  hierarchies for robust visual tracking. CoRR abs/1501.04587

\bibitem[{Wang et~al(2016)Wang, Zhang, Liu, and
  Metaxas}]{wang_reliable_memories}
Wang S, Zhang S, Liu W, Metaxas DN (2016) Visual tracking with reliable
  memories. In: Proceedings of the Twenty-Fifth International Joint Conference
  on Artificial Intelligence, pp 3491--3497

\bibitem[{Wei~Zhong(2012)}]{scm_cvpr2012}
Wei~Zhong MHY Huchuan~Lu (2012) Robust object tracking via sparsity-based
  collaborative model. In: Comp. Vis. Patt. Recognition, pp 1838--1845

\bibitem[{van~de Weijer et~al(2009)van~de Weijer, Schmid, Verbeek, and
  Larlus}]{colornames_tip2009}
van~de Weijer J, Schmid C, Verbeek J, Larlus D (2009) Learning color names for
  real-world applications. IEEE Trans Image Proc 18(7):1512--1523

\bibitem[{Wu et~al(2013)Wu, Lim, and Yang}]{otb_cvpr2010}
Wu Y, Lim J, Yang MH (2013) Online object tracking: A benchmark. In: Comp. Vis.
  Patt. Recognition, pp 2411-- 2418

\bibitem[{Wu et~al(2015)Wu, Lim, and Yang}]{otb_pami2015}
Wu Y, Lim J, Yang MH (2015) Object tracking benchmark. IEEE Trans Pattern Anal
  Mach Intell 37(9):1834--1848

\bibitem[{Xu~Jia(2012)}]{asla_cvpr2012}
Xu~Jia MHY Huchuan~Lu (2012) Visual tracking via adaptive structural local
  sparse appearance model. In: Comp. Vis. Patt. Recognition, pp 1822--1829

\bibitem[{Zhang et~al(2014)Zhang, Zhang, Liu, Zhang, and
  Yang}]{zhang_stc_eccv2014}
Zhang K, Zhang L, Liu Q, Zhang D, Yang MH (2014) Fast visual tracking via dense
  spatio-temporal context learning. In: Proc. European Conf. Computer Vision,
  Springer International Publishing, pp 127--141

\bibitem[{Zhu et~al(2016)Zhu, Porikli, and Li}]{ebt_cvpr2016}
Zhu G, Porikli F, Li H (2016) Beyond local search: Tracking objects everywhere
  with instance-specific proposals. In: The IEEE Conference on Computer Vision
  and Pattern Recognition (CVPR), pp 943--951

\end{thebibliography}

\end{document}